\documentclass[lettersize,journal]{IEEEtran}
\usepackage{cite}
\usepackage{amsmath, amsfonts}
\usepackage{algorithmic}
\usepackage{array}
\usepackage[caption=false,font=normalsize,labelfont=sf,textfont=sf]{subfig}
\usepackage{textcomp}
\usepackage{stfloats}
\usepackage{url}
\usepackage{verbatim}
\usepackage{graphicx}
\def\BibTeX{{\rm B\kern-.05em{\sc i\kern-.025em b}\kern-.08em
		T\kern-.1667em\lower.7ex\hbox{E}\kern-.125emX}}
\usepackage{balance}

\usepackage{soul} 
\usepackage{color, xcolor} 
\sethlcolor{pink}
\soulregister\cite7 
\soulregister\ref7 
\usepackage{multirow}
\usepackage{makecell}

\usepackage{enumitem}
\usepackage[misc]{ifsym} 
\usepackage{threeparttable}
\usepackage{booktabs}
\usepackage{tabularx}
\usepackage{amsmath}

\usepackage{bm} 

\begin{document}

\title{	
Decentralized Heterogeneous Multi-Robot Collaborative Exploration for Indoor and Outdoor 3D Environments
}

\author{
 Yuxiang Li, Kun Chen, Jiancheng Wang, Shihao Fang, 
 Haoyao Chen*, \emph{Senior Member, IEEE}, 
 Yunhui Liu, \emph{Fellow, IEEE}
\thanks{
	This work was supported in part by xxxxxx
	
	Yuxiang Li, Kun Chen, Jiancheng Wang, Shihao Fang, and Haoyao Chen are with the College of Artificial Intelligence, 
	Harbin Institute of Technology (Shenzhen), Shenzhen 518055, China, 
	e-mail: (\{19B953004,  
	210330122,  
	22B353018,  
	25B965002,  
	\} @stu.hit.edu.cn, hychen5@hit.edu.cn).
	
	Yunhui Liu is with the Department of Mechanical and Automation Engineering, The Chinese University of Hong Kong, Hongkong 999077, China, e-mail: (yhliu@cuhk.edu.hk).
}
}


\maketitle

\begin{abstract}
Heterogeneous multi-robot systems feature significant adaptability for complex environments. However, effective collaboration that fully exploits the robots' potential remains a core challenge. This paper proposes a decentralized collaborative framework for heterogeneous multi-robot systems to autonomously explore indoor and outdoor 3D environments. 
First, a basic perception map that integrates terrain and observation metrics is designed. Improved supervoxel segmentation is developed to simplify the map structure and form a high-level representation that supports lightweight communication. Second, the traversal and observation capabilities of heterogeneous robots are modeled to evaluate the requirements of task views derived from incomplete supervoxels. These task views are grouped by requirements and clustered to streamline assignment. 
Subsequently, the view-cluster assignment is formulated as a heterogeneous multi-depot multi-traveling salesman problem (HMDMTSP) that incorporates constraints between view-cluster requirements and robot capabilities. An improved genetic algorithm is developed to efficiently solve this problem while ensuring global consistency. 
Based on the assignments, redundant views within clusters are eliminated to refine exploration routes. 
Finally, conflicts between robots' motion paths are resolved. Simulations and field experiments in cluttered indoor and outdoor environments demonstrate that our approach effectively coordinates exploration tasks among heterogeneous robots, achieving superior exploration efficiency and communication savings compared to state-of-the-art approaches.


\end{abstract}


\section{Introduction} \label{sec:introduction}
\IEEEPARstart{A}{utonomous} exploration of unknown environments is a fundamental task for robotic systems, widely applied in areas such as information gathering \cite{cpp}, 
disaster response \cite{tian2020search}, and source searching \cite{source_searching}. Real-world environments pose significant challenges for multi-robot collaborative exploration, particularly in mixed indoor-outdoor regions with complex layouts, confined regions, and uneven terrains. These scenes demand robust real-time perception, path planning, and collaborative decision-making capabilities from robots. Heterogeneous multi-robot systems, with diverse configurations and functionalities, offer great potential for adapting to such complex environments. However, the heterogeneity complicates exploration planning, requiring coordinated map maintenance, efficient information sharing, optimized task assignment, and consistent collaborative strategies.


Existing multi-robot exploration approaches are primarily designed for homogeneous systems and can be broadly categorized into two types: fine-grained reconstruction-oriented approaches and communication-constrained task-oriented approaches. Fine-grained reconstruction-oriented approaches \cite{dong2019multi, hardouin2023multirobot, bartolomei2023fast} generate high-precision environmental models by sharing raw map data among robots for view planning, thereby enabling the capture of geometric details. However, these approaches rely heavily on centralized interaction with large volumes of map data, imposing extremely high demands on real-time, high-reliability communication. 
Communication-constrained task-oriented approaches \cite{2023RACER, yan2022mui, dong2024fast} focus on environmental abstraction—such as spatial simplification or geometric approximation—to allocate exploration regions and reduce communication overhead among robots. However, these approaches face limitations in unstructured or complex three-dimensional environments. Moreover, the assumption of robot homogeneity hinders the potential of heterogeneous multi-robot systems.

Recent studies have proposed heterogeneous multi-robot exploration approaches, which can be broadly categorized into synchronous and asynchronous collaboration. Synchronous multi-robot exploration requires robots to execute tasks simultaneously \cite{tranzatto2022cerberus, morrell2022nebula, chen2021clustering}. Mainstream approaches enable collaboration through either hierarchical planning frameworks \cite{tranzatto2022cerberus} or topology-driven strategies \cite{morrell2022nebula}. Although guiding exploration with centrally communicated global information can reduce transmission overhead, the lack of in-depth consideration of local reconstruction quality and task conflicts among robots limits its applicability.
Asynchronous multi-robot exploration allows tasks to be executed sequentially across robots, typically focusing on one-to-one collaboration strategies\cite{yue2021tightly} and dynamic triggering mechanisms\cite{qin2019autonomous}. For example, when one robot encounters difficulties during exploration, another robot takes over to continue the task \cite{yue2021tightly}. Alternatively, in a coarse-to-fine manner, one robot initially performs a broad exploration, guiding others to conduct more detailed scanning \cite{qin2019autonomous}. However, such operations inevitably prolong the overall exploration duration. Furthermore, one-to-one collaboration strategies may be insufficient for many-to-many robot cooperation, especially in complex scenes involving rugged terrains, obstructed structures, or transitions between indoor and outdoor regions.

This paper focuses on three major challenges in heterogeneous multi-robot exploration: balancing fine-grained mapping with communication efficiency, accounting for robot-capability differences in view planning, and coordinating multiple robots in complex 3D environments. To this end, we propose a decentralized collaborative exploration planning framework that significantly improves mapping efficiency and reconstruction completeness, while substantially reducing communication overhead in heterogeneous multi-robot systems. 
First, a basic perception map that integrates terrain and observation metrics is proposed, and improved supervoxel segmentation is developed to simplify the map. Then, the generated supervoxels and their connectivity relationships are utilized to construct a high-level representation, enabling efficient information sharing and low-bandwidth communication among robots. Afterward, task views are generated from supervoxels, and the traversal and observation capabilities of heterogeneous robots are modeled. Task views are further grouped by requirements and clustered to streamline assignment. Subsequently, the assignment problem is formulated, and an improved genetic algorithm is developed for efficient optimization, incorporating requirement–capability matching constraints. Finally, according to the assignments, efficient exploration routes are generated for each robot, 
with strategies designed to resolve conflicts for executable motion paths.
The main contributions of this paper are as follows:

\begin{enumerate}
\item 
A basic perception mapping method for complex environments is proposed, incorporating multiple metrics to evaluate traversability and coverage status, thereby overcoming the limitations of traditional maps in heterogeneous multi-robot systems.

\item 
A high-level abstract map representation is developed based on the improved supervoxels and their connectivity relationships, enabling globally consistent map fusion and low-bandwidth communication among robots.

\item 
A novel view-planning method is proposed, which models the mobility and observation capabilities of distinct robot species and evaluates task-view requirements to guide heterogeneous multi-robot collaborative exploration. 

\item 
The task assignment is modeled as an HMDMTSP with requirement–capability matching constraints. An improved genetic algorithm is designed to accelerate the solving process, with redefined genetic representation and fitness function.
The project will be open-sourced on GitHub\footnote{
https://github.com/xxxxx-xxxx/xxxxxx-xxxxxx-xxx.git
}.
\end{enumerate}

\section{Related Work} \label{sec:related-work}
Studies on multi-robot collaborative exploration can be categorized into homogeneous and heterogeneous approaches.

\subsection{Homogeneous Multi-Robot Collaborative Exploration}
\textbf{Fine-grained reconstruction-oriented approaches} focus on detailed surface scanning and share raw maps among robots. Dong et al. \cite{dong2019multi} propose a multi-robot collaborative 3D reconstruction approach that constructs an uncertainty map to quantify insufficiently scanned regions and discretizes the view space to select effective views. 
Hardouin et al. \cite{hardouin2023multirobot} extract incomplete surface elements to generate task views and propose a traveling salesman–greedy exploration strategy for allocating view clusters among robots. 
Bartolomei et al. \cite{bartolomei2023fast} propose a decentralized multi-robot exploration approach that generates task views based on frontier clusters  
and alternates between exploration and collection modes to fully cover isolated frontier clusters left behind, thereby avoiding long detours. 
\textcolor{black}
{Asgharivaskasi et al. \cite{riemannian} integrate communication, mapping, and exploration planning into a Riemannian-optimization-based framework for distributed active mapping.}
Although these approaches enable effective multi-robot cooperative 3D reconstruction, they rely heavily on extensive map exchange, making them difficult to apply in communication-constrained scenes. To this end, we introduce observation metrics into the basic perception map to evaluate reconstruction quality, while constructing a high-level map representation to reduce communication overhead. Different from the approaches \cite{dong2019multi, hardouin2023multirobot, bartolomei2023fast, riemannian}, we propose a decentralized framework for map maintenance and exploration planning to ensure consistency in 3D reconstruction and task assignment across multiple robots.

\textbf{Communication-constrained task-oriented approaches} emphasize global spatial abstraction and partitioning to reduce communication data and optimize exploration-region assignment. Meanwhile, their local exploration strategies focus on improving reconstruction quality. 
Zhou et al. \cite{2023RACER} and Yan et al. \cite{yan2022mui} achieve collaborative exploration by partitioning the workspace into subspaces, generating global coverage paths through subspace traversal, and refining them locally within each robot’s subspace.
Dong et al. \cite{bartolomei2023fast} partition the 3D exploration space into uniform grid nodes, construct a topology graph for inter-robot sharing, and allocate nodes using dynamic voronoi partitioning. 
However, spatial gridding may erroneously split connected regions, resulting in unnecessary detours during exploration. An alternative approach approximates regions with disks or star-convex polyhedra, thereby reducing communication overhead and preventing overlap in exploration areas. 
Kim et al. \cite{kim2023multi} employed disk fitting for room recognition to guide collaborative indoor exploration, where robots share only disk information.  
Gao et al. \cite{gao2022meeting} employ star-convex polyhedra to represent known free space for lightweight communication. 
Renzaglia et al. \cite{renzaglia2020common} and Bayer et al. \cite{bayer2021decentralized} address underground cave exploration by constructing topology graphs, where nodes are evaluated based on information gain and assigned to robots using either greedy or directed-tree strategies. 
\textcolor{black}
{Zhao et al. \cite{zhao2025multirobot} incorporate reinforcement learning into multi-robot exploration to improve efficiency, which requires extensive training on 2D maps.} 
Additionally, such approaches are limited by 2D spatial simplifications or geometric approximations, making them difficult to apply in irregular environments or complex 3D structures. To this end, we introduce a supervoxel-based map representation that retains key environmental features for effective view generation and assignment in complex settings while maintaining lightweight communication.

\subsection{Heterogeneous Multi-Robot Collaborative Exploration}
\textbf{Synchronous multi-robot collaborative exploration approaches} require robots to execute tasks simultaneously. Tranzatto et al. \cite{tranzatto2022cerberus} present a graph-based aerial–ground collaboration framework that utilizes node information gain to guide exploration planning. 
Similarly, Agha et al. \cite{morrell2022nebula} design a belief-space representation and a hierarchical coverage-planning framework that uses an informative roadmap to collaborate heterogeneous robots. However, these approaches achieve centralized multi-robot collaboration by leveraging graph structures tailored to subterranean-topology exploration, which limits their applicability to other environments, such as indoor or outdoor scenes. 
Chen et al. \cite{chen2021clustering} address the cooperative coverage problem for heterogeneous aerial robots by integrating indicators such as flight altitude and sensing range. 
Liu et al. \cite{liu2023active} incorporate semantic information to maintain consistency of multi-robot maps while reducing communication overhead.  
Zhang et al. \cite{zheng2025aage} propose a collaborative aerial–ground exploration framework in which the aerial robot generates bird’s-eye views to identify key regions that guide the ground robot efficiently. 
Li et al. \cite{li2021energy} introduce a multi-task Gaussian process approach that fuses images from an aerial robot with point clouds from a ground robot to construct traversability maps and guide both robots in exploration planning. 
These approaches have demonstrated effective heterogeneous multi-robot collaboration in specific environments and tasks. 
However, they prioritize rapid volumetric coverage or predefined target scanning, neglecting reconstruction quality in execution and failing to address overlapping/conflicting tasks. 
To this end, we formulate task views with explicit capability requirements for heterogeneous multi-robot assignment, eliminate redundancy through co-visibility analysis, and resolve inter-robot conflicts, thereby improving collaboration efficiency.

\textbf{Asynchronous multi-robot collaborative exploration approaches}  require robots to execute tasks alternately. 
Qin et al. \cite{qin2019autonomous} propose a dual-layer exploration strategy for aerial and ground robots, in which the ground robot is first tasked with fast scanning, and the generated rough map is then used to guide the aerial robot in performing fine mapping. 
Yue et al. \cite{yue2021tightly} propose a high-level collaboration strategy in which a ground robot is equipped with an aerial robot to perform exploration tasks. A dynamic event-triggering mechanism is designed: when the ground robot encounters an intractable obstacle, the aerial robot is triggered to take off and continue the task. 
Zhang et al. \cite{zhang2022fast} and Li et al. \cite{li2024enhancing} propose similar aerial-ground collaborative exploration approaches, where the aerial robot leads the exploration and constructs a traversability map for the ground robot to plan feasible paths. 
These approaches focus on one-to-one collaboration in exploration tasks, where one robot assists another in executing tasks sequentially. However, their application to one-to-many or many-to-many heterogeneous multi-robot collaborations could be limited. 
To address these issues, this paper incorporates robot-capability analysis into view generation, task assignment, and view planning to maximize system performance of heterogeneous ground-robot teams.

\begin{figure*}[ht]
\centering
\includegraphics[width=1\linewidth]{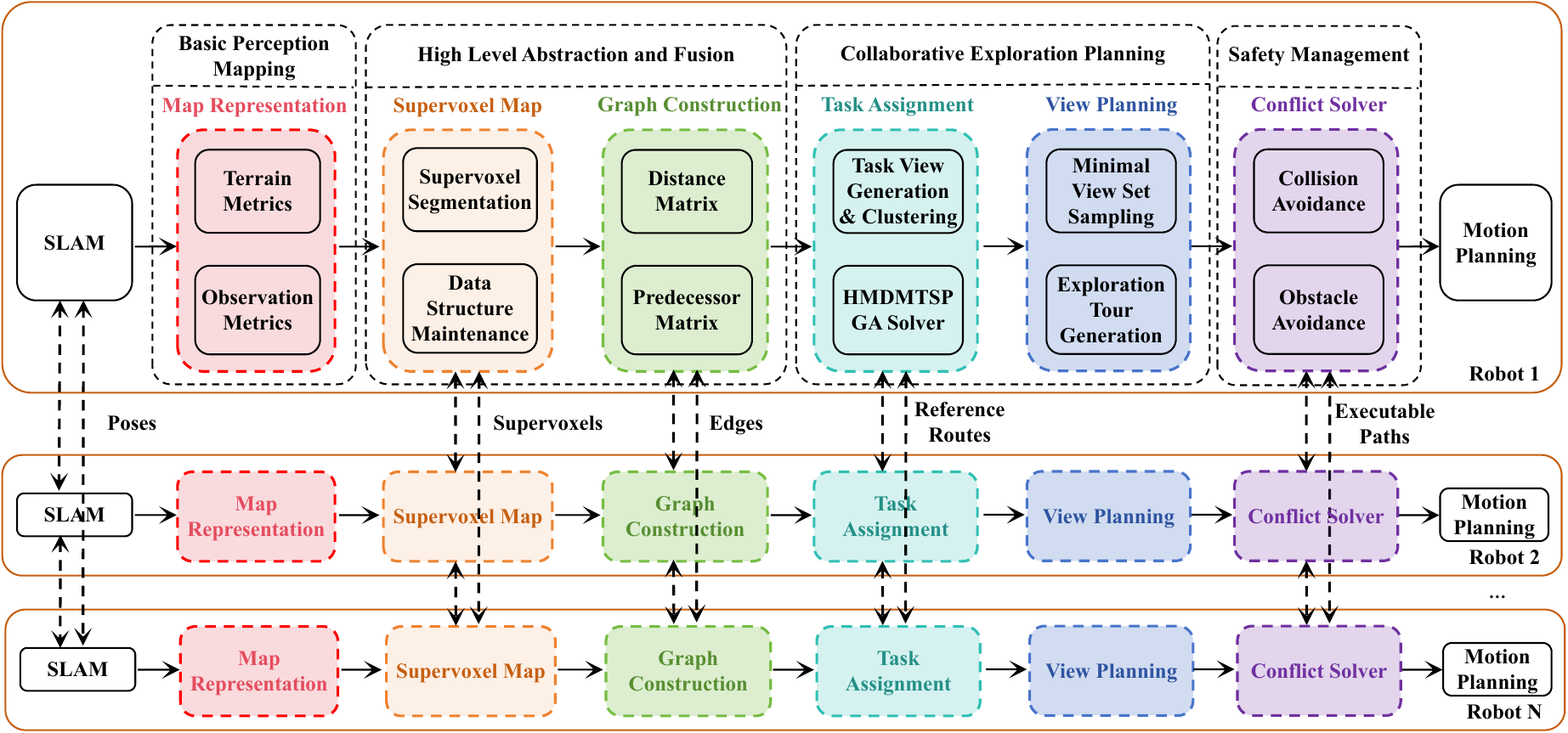}
\caption{
	The proposed decentralized heterogeneous multi-robot collaborative exploration framework for \textcolor{black}{ indoor/outdoor/mixed} 3D environments. 
	The framework primarily comprises basic perception mapping, high-level abstraction and fusion, collaborative exploration planning, and safety management. 
    \textcolor{black}{ Dashed arrows represent the information shared among robots. }
}
\label{fig:overview}
\end{figure*}

\section{System Overview}  \label{sec:system-overview}
The proposed framework aims to collaboratively explore an entirely unknown environment with a heterogeneous multi-robot team in a decentralized manner (see Fig. \ref{fig:overview}). The team comprises multiple robot species, each with distinct observation and mobility capabilities. The robots independently perform simultaneous localization and mapping (SLAM), sharing point clouds only for the initial map registration and continuously exchanging poses for subsequent mapping and exploration planning. 
Based on the terrain-aware mapping \cite{mytro}, this paper not only adopts terrain metrics to assess the traversability of different robot species but also introduces observation metrics to indicate coverage status, forming the basic perception map for each robot (see Sec. \ref{sec:map_representation_fusion}). 
To simplify the map representation, we perform supervoxel segmentation (see Sec. \ref{sec:supervoxel_map}) and form a supervoxel map. Meanwhile, a traversal-topology graph (see Sec. \ref{sec:topological_map}) is constructed, in which traversable supervoxels serve as nodes and the connectivities between them as edges. Distance and predecessor matrices are maintained for the graph to enable efficient distance queries and route searches for subsequent task assignment and view planning.  
The supervoxels and topology graph constitute a high-level map, incrementally transmitted within the robot team to enable lightweight communication and global consistency. (see Sec. \ref{sec:map_integration}). 

Subsequently, task views are generated from the supervoxel map (see Sec. \ref{sec:task_view_generation}) and clustered by their capability requirements (see Sec. \ref{sec:view_clustering}). 
The assignment of task clusters is modeled as a HMDMTSP, with the matching between robot species and task-cluster requirements incorporated as constraints. 
An efficient genetic algorithm is developed to solve the traversal sequence of task clusters for multiple robots, with robots exchanging their sequences to enhance global solution consistency (see Sec. \ref{sec:hdmtsp}). 
Then, the minimal view set sampling (MNSS) approach \cite{mytim} is used to streamline the task views within each cluster, yielding efficient exploration routes (see Sec. \ref{sec:exploration_tour_generation}). 
Each robot sequentially navigates to the task views along the assigned exploration route by generating executable paths using motion planning approaches such as \textcolor{black}{the hierarchical planner} \cite{mytro}. The executable paths are shared within the team to resolve conflicts, such as collision and obstacle avoidance, for safety management (see Sec. \ref{sec:resolve_path_conflicts}). 
After several rounds of collaborative planning, the exploration is progressively completed, ultimately achieving a 3D reconstruction of the entire environment. 
Without loss of generality, vision- or LiDAR-based object detection methods \cite{yolo, rangenet} enable robots to detect one another within their respective obervation ranges, facilitating obstacle avoidance and dynamic point-cloud filtering. 
In such a case, real-time pose transmission can be omitted to further reduce communication overhead. 

\section{Multi-Robot Map Representation and Fusion} \label{sec:map_representation}
The mapping process should explicitly consider differences in mobility and sensing capabilities across robot species to support heterogeneous robot teams. To this end, we propose a novel map representation that evaluates the traversability of different robot species using terrain metrics, while incorporating observation metrics to indicate the coverage status. 
The map is further simplified via improved supervoxel segmentation to enable low-bandwidth sharing among heterogeneous robots, while maintaining globally consistent information to support task assignment and view planning. 

\subsection{Basic Perception-Map Representation} \label{sec:map_representation_fusion}
\subsubsection{Terrain and Observation Metrics}
The basic perception-map representation is based on the terrain-aware map representation \cite{mytro}, which is applicable to various ground-robot species and diverse terrain conditions. 
The map is represented using an octree-based data structure to manage voxels, each of which encodes a Gaussian distribution. For  
each occupied voxel $v\in\bm{V}$ in the map, 
terrain metrics, such as surface normals  $\bm{n}(v)$, slope $s(v)$, and sparsity $d(v)$, are estimated by fusing neighboring voxels' Gaussian distributions. These metrics are then utilized to evaluate the boolean traversal risk $\mathcal{R}(v)$. 
For decentralized collaborating heterogeneous robots, traversability should be evaluated according to each robot species’s capabilities, with different weights and thresholds reflecting their specific mobility characteristics. 
Rather than binarizing traversal costs as in \cite{mytro}, we reformulate traversability by combining traversal risks from multiple robot species, thereby ensuring computational simplicity.
Let the robot team be defined as  $\bm{R}=\{{r_{1},r_{2}, ...,r_{N_{r}}}\}$, containing $N_r$ robots.
The team includes $N_k$ robot species $k\in \bm{K}$, forming the set $\bm{K}=\{{k_{1},k_{2}, ...,k_{N_{k}}}\}$.
Each robot $r\in\bm{R}$  is associated with a species through the mapping $f_{k}(r): \bm{R} \longmapsto \bm{K}$. Accordingly, the boolean traversability $\zeta(v)$ is defined as follows:

\begin{equation}
\label{eq4-1}
\zeta(v)=
\begin{cases}
	1,   \quad \text{if~} \exists k\in \bm{K},\mathcal{R}_k(v)=0 \\
	0,   \quad \text{otherwise}  
\end{cases},
\end{equation} \textcolor{black}{
where a voxel is defined as traversable if it is treated as low traversal risk by at least one robot species. }

\begin{figure}[tpb]
\centering
\includegraphics[width=0.6\linewidth]{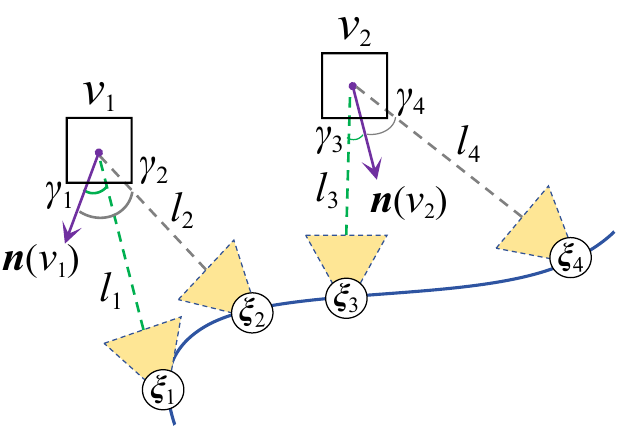}
\caption{
	The voxel-view relationship from a top-down view. 
}
\label{fig:observation-metric}
\end{figure}

Real-world scenes are typically complex, and their irregular surfaces and self-occlusions often cause incomplete scanning and map reconstruction. 
To address these challenges, appropriate observation metrics are required to evaluate reconstruction quality and to guide task-view generation. Specifically, not only is sufficient point-cloud density required to accurately approximate object surfaces using Gaussian distributions, but also suitable observation positions and directions are essential to reduce measurement uncertainty from depth sensors.
Therefore, we introduce observation density $\rho(v)$, observation angle $\gamma(v)$, and observation distance $l(v)$ as evaluation metrics of reconstruction quality and store them as voxel attributes in the map. 
Observation density represents the number of measurement points per unit area and is inversely proportional to the aforementioned sparsity in terrain metrics. The observation angle is defined as the minimum angle between the inverse view direction and the normal vector of the observed voxel, while the observation distance corresponds to the minimum distance between the sensor and the voxel. 
As shown in Fig. \ref{fig:observation-metric}, the top-down view illustrates the relationship between voxels and the views along the exploration route (blue curve). The purple arrows represent the normal vectors of the voxels. 
Voxel $v_{1}$ is observed by views $\bm{\xi}_{1}$ and $\bm{\xi}_{2}$, while $v_{2}$ is observed by views $\bm{\xi}_{3}$ and $\bm{\xi}_{4}$. 
For views $\bm{\xi}_{1}$ and $\bm{\xi}_{2}$, the observation distances  $l_1$ and $l_2$ both lie within the threshold $l_{th}$, whereas the observation angle $\gamma_2$ exceeds the threshold $\gamma_{th}$; thus, \textcolor{black}{$\gamma(v_1)=\gamma_{1}$, $l(v_1)=l_{1}$. }
For views $\bm{\xi}_{3}$ and $\bm{\xi}_{4}$, both the observation angles and distances lie within the thresholds; therefore,  \textcolor{black}{$\gamma(v_2)=\operatorname*{min}\left(\gamma_{3},\gamma_{4}\right)$, $l(v_2)=\operatorname*{min}\left(l_{3},l_{4}\right)$. }

Then, the coverage status $Q(v)$ is introduced to indicate whether the voxel $v$ has been adequately observed, and the expression is defined as follows: 
\begin{equation}
\label{eq2}
Q(v)=(\rho(v)>\rho_{th})\wedge(\gamma(v)<\gamma_{th})\wedge(l(v)<l_{th}),
\end{equation}
where $\wedge$ denotes the logical AND operator, and $\rho_{th}$ is the observation-density threshold. 
Voxels that satisfy the above conditions are defined as Complete Surface Elements (CSE) and the others as Incomplete Surface Elements (ISE). 
The aforementioned terrain and observation metrics are stored as voxel attributes in the map and are updated during the point cloud integration. 

\subsubsection{Voxel Set Partitioning}
Since the motion space of ground robots is constrained to traversable areas, exploration planning should consider both the view-ISE visibility and the view-position traversability. 
Accordingly, exploration planning seems to cover the object surfaces while acquiring sufficient terrain observations for generating feasible paths. 
To this end, the map voxels $\bm{V}$ are further partitioned using terrain and observation metrics. First, based on the slope thresold $s_{th}$, the voxels are partitioned into the facade voxel set $\bm{U}$ and the ground voxel set $\bm{G}$:
\begin{equation}
\label{eq3}
\begin{cases}
	\bm{U}=\{v\mid s(v)\geq s_{th},v\in \bm{\bm{V}}\} \\
	\bm{G}=\{v\mid s(v)<s_{th},v\in \bm{\bm{V}}\}
\end{cases}.
\end{equation}
Based on the coverage status defined in formula (\ref{eq2}), the voxel sets are further divided into the facade CSE subset $\bm{U}_C$, the facade-ISE subset $\bm{U}_I$, the ground CSE subset $\bm{G}_C$, and the ground-ISE subset $\bm{G}_I$. 
Meanwhile, the ground voxel set needs to be subdivided according to the traversability defined in formula (\ref{eq4-1}). 
Since the traversable voxels form the motion space of ground robots, 
\textcolor{black}
{the ground CSE subset $\bm{G}_C$ and the ground-ISE subset $\bm{G}_I$ are obtained by partitioning only over traversable ground voxels. }
Therefore, the subsets are defined as follows: 

\begin{equation}
\label{eq5-3}
\begin{cases}

	\bm{U}_C=\{v\mid \textcolor{black}{Q(v)=1},v\in\bm{U}\} \\
	\bm{U}_I=\{v\mid \textcolor{black}{Q(v)=0},v\in\bm{U}\} \\
	\bm{G}_C=\{v\mid \textcolor{black}{Q(v)=1},\zeta(v)=1,v\in\bm{G}\} \\
	\bm{G}_I=\{v\mid \textcolor{black}{Q(v)=0},\zeta(v)=1,v\in\bm{G}\}

\end{cases}.
\end{equation}

The ISE subsets $\bm{U}_I$ and $\bm{G}_I$ serve as the coverage targets for exploration planning, while the ground subsets are used to generate candidate view positions. 
The partition enables views to be generated for different environmental surfaces, facilitating subsequent exploration planning. 

\begin{figure}[tpb]
\centering
\includegraphics[width=1\linewidth]{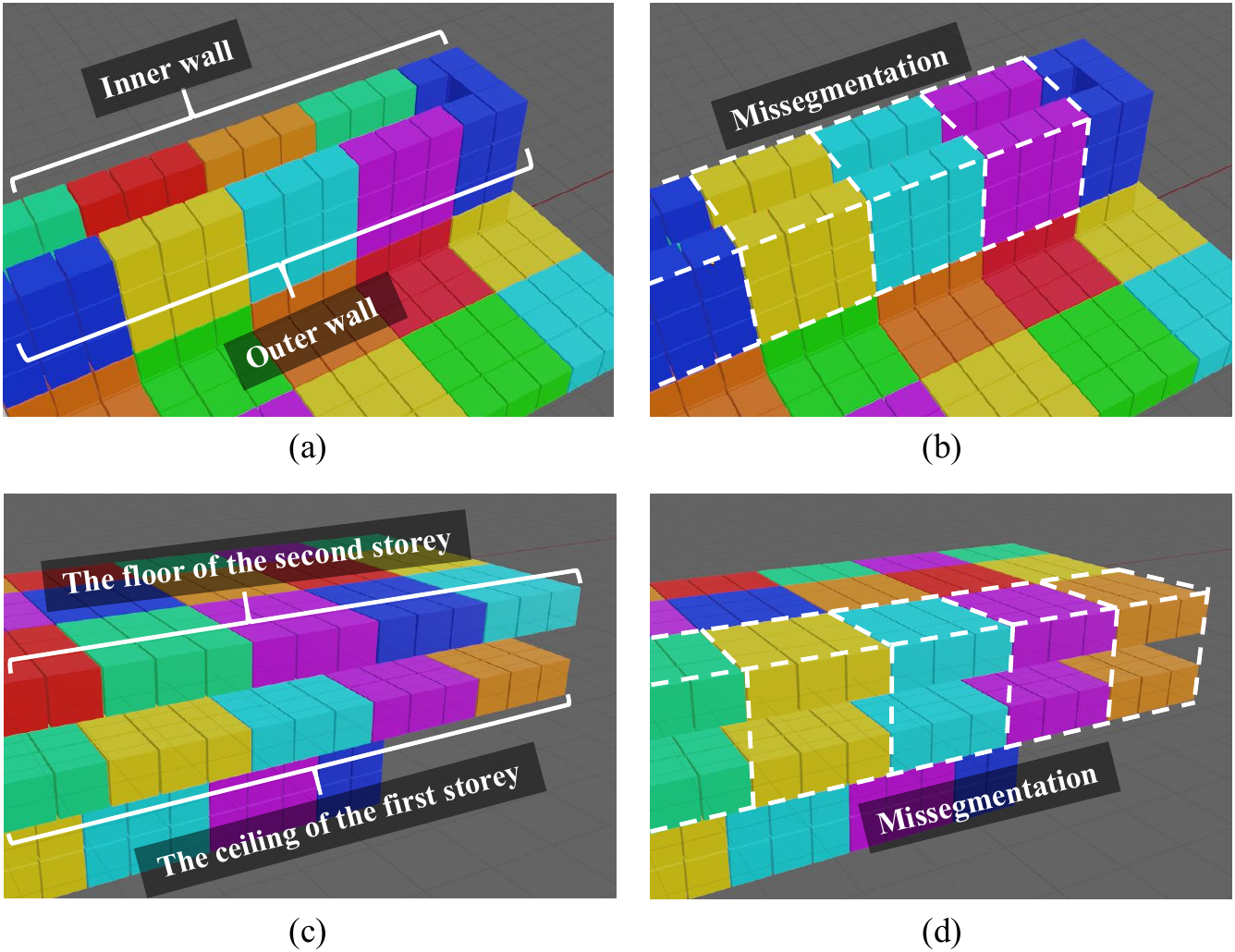}
\caption{
	Voxel segmentation results in two scenes. (a) Supervoxel segmentation of the inner and outer surfaces of the wall. (b) Uniform spatial division or Euclidean clustering of the inner and outer surfaces of the wall. (c) Supervoxel segmentation of the floor and ceiling. (d) Uniform spatial division or Euclidean clustering of the floor and ceiling. 
}
\label{fig:cluster-wall-ceil}
\end{figure}

\subsection{Supervoxel Map}  \label{sec:supervoxel_map}
\subsubsection{Supervoxel Segmentation}
Since numerous ground and facade voxels can be obtained in the map during exploration, generating views for each single ISE voxel is computationally intractable for real-time performance and leads to redundant routes. 
Moreover, transmitting raw voxel attribute data imposes a significant communication burden.  
Mainstream methods typically utilize uniform spatial division \cite{yan2022mui} or clustering-based \cite{hardouin2023multirobot} strategies to alleviate computation and communication overhead. 
However, uniform spatial division may confuse environmental boundaries (see Fig. \ref{fig:cluster-wall-ceil}(b) and (d)). Missegmentation of upper and lower floors, or indoor and outdoor spaces, leads to detours or incomplete reconstruction during exploration. 
Meanwhile, traditional clustering methods may yield inconsistent results when point clouds differ slightly, thereby degrading consistency across multiple robots. 
To overcome these issues, we propose a supervoxel-based high-level map representation. 

Supervoxel segmentation \cite{papon2013voxel} incorporates seed division and adjacency constraints into the clustering process, preserving spatial connectivity and compactness within each cluster (i.e., supervoxel).   
The generated clusters follow the 3D boundaries of environmental objects and are regularly shaped (see Fig. \ref{fig:cluster-wall-ceil}(a) and (c)). 
Moreover, voxelizing the space to generate seeds yields clusters that do not deviate significantly from the seed positions. Indexing the initial seeds enables consistent associations among high-level maps across multiple robots, thereby facilitating global map fusion. 
However, supervoxel segmentation remains insufficient for autonomous reconstruction. To address this, we introduce specific improvements in seed generation and connected region growth.  

\begin{figure}[tpb]
\centering
\includegraphics[width=1\linewidth]{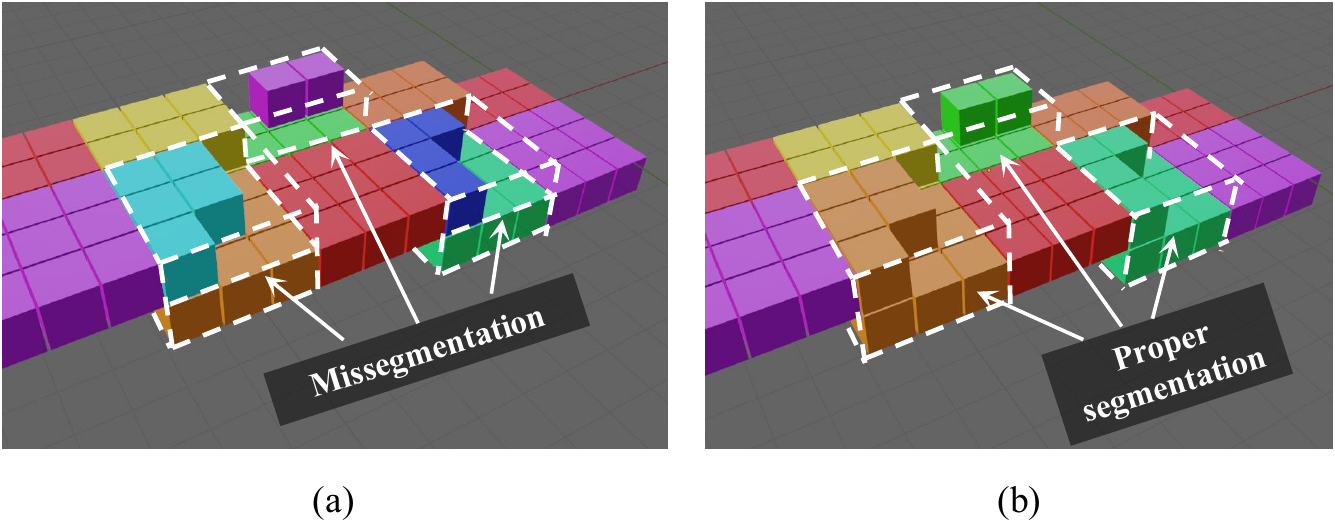}
\caption{
	Results of two supervoxel-segmentation methods for noisy ground voxels.
	(a) Results of the original supervoxel-segmentation method. 
	(b) Results of the improved supervoxel-segmentation method. 
}
\label{fig:cluster-noisy-ground}
\end{figure}

First, the original supervoxel segmentation voxelizes the space and selects the point closest to each supervoxel's center as an initial seed. Neighboring points are identified, and seeds with too few neighbors are removed to reduce noise.  
However, this approach is only suitable for foreground object segmentation in depth images and is not adaptable to LiDAR point clouds with large background regions. 
For example, ground point clouds captured by LiDAR suffer from measurement noise, resulting in thickened geometrical shapes that could be erroneously segmented into two layers (see Fig. \ref{fig:cluster-noisy-ground}(a)) and hindering exploration planning. 
To address this issue, we propose a breadth-first search (BFS)-based seed merging method. It applies BFS on the generated seeds and merges adjacent seeds within a distance threshold $\eta_e$ to prevent missegmentation caused by reverse region growing. 

Second, the original supervoxel segmentation partitions point clouds using an adjacency graph constructed by searching for neighboring points, with a KD-Tree\cite{kdtree} used to maintain the graph. This method is inefficient for continuously updated maps, as rebuilding the KD-Tree is computationally expensive. Therefore, we propose a wavefront-propagation \cite{wavefront}-based connected region-growing method that directly operates on the octree structure, efficiently expanding connected voxels from seeds. 
As shown in Fig. \ref{fig:cluster-noisy-ground}(b), the improved segmentation effectively prevents noisy ground regions from being split into multiple layers.

\begin{table*}[b]
\renewcommand\arraystretch{1.0} 
\centering
\caption{
	Example for Determining Supervoxel-Traversal Requirement
}  
\label{tab:4-1-supervoxel-traversal-level}  
\begin{tabular}{|cc|cccc|}
	\hline
	\multicolumn{1}{|c|}{\multirow{2}{*}{Robot Species $k$}} & \multirow{2}{*}{\begin{tabular}[c]{@{}c@{}}Robot Traversal \\ Capability $\overline{\zeta}(\hat{\bm{v}})$\end{tabular}} & \multicolumn{4}{c|}{Determined as Low Traversal Risk} \\ \cline{3-6} 
	\multicolumn{1}{|c|}{}  &  & \multicolumn{1}{c|}{Supervoxel 1} & \multicolumn{1}{c|}{Supervoxel 2} & \multicolumn{1}{c|}{Supervoxel 3} & Supervoxel 4 \\ \hline
	\multicolumn{1}{|c|}{A}  & Level 1  & \multicolumn{1}{c|}{No}  & \multicolumn{1}{c|}{No}   & \multicolumn{1}{c|}{No}  & Yes \\ \hline
	\multicolumn{1}{|c|}{B}  & Level 2  & \multicolumn{1}{c|}{No}  & \multicolumn{1}{c|}{No}   & \multicolumn{1}{c|}{Yes} & Yes \\ \hline
	\multicolumn{1}{|c|}{C}  & Level 3  & \multicolumn{1}{c|}{No}  & \multicolumn{1}{c|}{Yes}  & \multicolumn{1}{c|}{Yes} & Yes \\ \hline
	\multicolumn{2}{|c|}{Traversal Requirement of the Supervoxel}  & \multicolumn{1}{c|}{N/A}  & \multicolumn{1}{c|}{3}   & \multicolumn{1}{c|}{2} & 1 \\ \hline
\end{tabular}
\end{table*}

\subsubsection{Supervoxel Data Structure}\label{sec:supervoxelDataStructure}
Each robot maintains its own basic perception map and segments its supervoxels accordingly. These supervoxels are then shared among multiple robots and fused to form a unified global map. 
As the minimal unit of the high-level map, supervoxels must strike a balance between sufficient simplification to minimize communication overhead and adequate information preservation to maintain global consistency. 
To this end, we propose a hash-table-based supervoxel-maintenance method that generates seeds at a specified interval $\eta_s$ and constructs hash keys from the seeds' coordinates as indices. The hash keys and corresponding supervoxels’ attributes form ``key-value" pairs for efficient retrieval. 
Since all robots are unified to the world-coordinate frame during the initial map registration, the hash keys are globally consistent, and the hash table establishes global associations among robots. 

\textcolor{black}{
A supervoxel, i.e. a cluster of voxels, may have different terrain and observation metric values.} Since the voxels within the same cluster are adjacent to each other, their attribute values are relatively similar. Therefore, we further abstract the attributes for each supervoxel $\hat{\bm{v}}$, such as the center coordinate $\boldsymbol{p}(\hat{\bm{v}})$, normal vector $\boldsymbol{n}(\hat{\bm{v}})$, coverage status $Q(\hat{\bm{v}})$, and traversal requirement $\overline{\zeta}(\hat{\bm{v}})$. 
The center coordinate is taken as \textcolor{black}{the mean voxel coordinate} within the supervoxel. 
The mean normal vector of the voxels is approximated as the normal vector of the supervoxel.  
The supervoxel's coverage status depends on whether all voxels within the cluster have been covered: 
\begin{equation}
Q(\hat{\bm{v}})=
\begin{cases}
	1, & \text{if~}\forall v\in\hat{\bm{v}},\mathrm{~}Q(v)=1 \\
	0, & \text{otherwise}
\end{cases}. 
\end{equation}

Subsequently, the traversal requirement $\overline{\zeta}(\hat{\bm{v}})$  of a supervoxel $\hat{\bm{v}}$ is determined based on the traversal capabilities of robot species. First, the traversal capabilities of various robot species $k \in \bm{K}$ are  sorted in ascending order, assigning each species $k$ a traversal-capability level $a_{\zeta}(k)\in\{1,2,...,N_{k}\}$. 
Then, the supervoxel's traversal requirement $\overline{\zeta}(\hat{\bm{v}})$ is defined as the minimum traversal-capability level required for all constituent voxels $v\in\hat{\bm{v}}$ to exhibit low traversal risk: 
\begin{equation}
\overline{\zeta}(\hat{\bm{v}})=\min\left\{a_\zeta(k),k\in \bm{K}\mid\forall v\in\hat{\bm{v}},\mathcal{R}_k(v)=0\right\}.
\end{equation}
An example of determining the traversal requirements of supervoxels is shown in Table \ref{tab:4-1-supervoxel-traversal-level}. 
A higher traversal requirement of a supervoxel corresponds to higher terrain difficulty that demands greater robot-traversal capability. 

Extracting supervoxels with the voxel subsets in formula (\ref{eq5-3}) produces the corresponding  
ground-CSE supervoxel set $\widehat{\boldsymbol{G}}_{C}$, 
ground-ISE supervoxel set $\widehat{\boldsymbol{G}}_{I}$, 
facade-CSE supervoxel set $\widehat{\boldsymbol{U}}_{C}$, and 
facade-ISE supervoxel set $\widehat{\boldsymbol{U}}_{I}$. 
By merging these sets, a facade supervoxel set $\widehat{\boldsymbol{U}}=\widehat{\boldsymbol{U}}_{C}\cup\widehat{\boldsymbol{U}}_{I}$ and a ground supervoxel set $\widehat{\boldsymbol{G}}=\widehat{\boldsymbol{G}}_{C}\cup\widehat{\boldsymbol{G}}_{I}$ are obtained for subsequent topology-map construction and task-view generation. 

\subsection{Traversal-Topology Graph}  \label{sec:topological_map}
We further incorporate a topology map  $\bm{\mathcal{G}}=(\bm{\mathcal{V}},\bm{\mathcal{E}})$ of the traversable ground areas into the high-level map to accelerate exploration planning. 
The vertex set $\bm{\mathcal{V}}=\{\hat{\bm{v}}|\hat{\bm{v}}\in\bm{\widehat{G}}\}$ consists of traversable ground supervoxels, while the edge set $\bm{\mathcal{E}}$ comprises connections between spatially adjacent supervoxels, with weights given by Euclidean distance.
As the number of supervoxel nodes increases with map expansion during exploration, graph-based pathfinding algorithms such as A* \cite{cui2011based} become more computationally expensive. 
To ensure real-time computation, a weighted adjacency matrix $\boldsymbol{\Pi}$ is first constructed for the graph, using the supervoxel nodes and their interconnecting edges. 
Subsequently, the GPU-accelerated Floyd-Warshall algorithm \cite{smith2010multi} is employed to quickly compute the distance matrix $\bm{\mathcal{D}}$ and the predecessor matrix $\bm{\mathcal{P}}$ from the weighted adjacency matrix. The dimensions of the matrices  $\bm{\mathcal{D}}$, $\boldsymbol{\Pi}$, and $\bm{\mathcal{P}}$ are  $N_{g}\times N_{g}\times N_{k}$, where $N_g=|\bm{\mathcal{V}}|$. 
The distance matrix $\bm{\mathcal{D}}$ records the graph distance between any two reachable nodes, while the predecessor matrix $\bm{\mathcal{P}}$ stores the index of the intermediate node along the shortest route between any two nodes.  
Note that the construction of adjacency matrix should respect the supervoxel traversal requirements and robot-species capabilities. If the traversal requirement of a supervoxel exceeds the robot-species capabilities, the corresponding entry in the matrix is set to infinity.  
Here, the function $\text{dist}_{\bm{\mathcal{G}}}(\hat{\bm{v}}_{i},\hat{\bm{v}}_{j},k)$ queries the distance between two nodes $\hat{\bm{v}}_i,\hat{\bm{v}}_j\in\bm{\mathcal{V}}$ on the graph  $\bm{\mathcal{G}}$, given the robot species $k\in \bm{K}$.  The function $\text{path}_{\bm{\mathcal{G}}}(\hat{\bm{v}}_i,\hat{\bm{v}}_j,k)$ queries feasible routes on the topology graph, and the function $\text{neigh}_{\bm{\mathcal{G}}}(\bm{p})$ identifies the nearest neighbor nodes to the position $\bm{p}$. 

\subsection{Incremental Transmission and Map Fusion}  \label{sec:map_integration}
The supervoxels and traversal-topology graph constitute a high-level map representation. 
Since each sensor frame covers a limited area and there is substantial overlap between consecutive frames, it is unnecessary to frequently broadcast the entire high-level map of each robot. 
Therefore, we design an incremental data-transmission approach, publishing only the updated supervoxels and edges within a specific time window  $\eta_c$. 
This significantly reduces redundant data transmission while maintaining the timeliness of data exchange. Subsequently, data is serialized in a compact form at the sender and deserialized at the receiver, avoiding the extra computational cost and distortion caused by data compression. 
Once a robot receives data from others, it updates its supervoxel map and topology graph to maintain a high-level map. 
The attributes of the supervoxels are fused to maintain global consistency across robots. 
Specifically, \textcolor{black}{based on the overlapping supervoxels produced by multiple robots, the traversal requirement of a supervoxel is determined by the maximum value; the coverage state is computed using a logical OR operation; and the normal vector is calculated as the average. 
}
In practice, the synchronizing time window $\eta_c$ is determined based on the communication bandwidth, preventing network congestion from excessively long windows while ensuring timely global synchronization before the next planning cycle. 

\section{Heterogeneous Multi-Robot Collaborative Exploration Planning} 
\label{sec:exploration_planning} 
Based on the globally maintained high-level map, we further propose methods for task-view generation and assignment as well as for exploration tour planning. Task views for ISE supervoxels are generated based on robot-species capabilities and classified as public or private categories. 
Then, the views are clustered to simplify task assignment.
The assignment problem is formulated as HMDMTSP and solved by improving the metaheuristic algorithm in \cite{miloradovic2021gmp}. Efficient exploration routes are then generated via minimal view-set sampling, and a conflict-resolution strategy is designed for safety management. 

\subsection{Task-View Generation} \label{sec:task_view_generation}
\begin{figure}[tpb]
\centering
\includegraphics[width=0.6\linewidth]{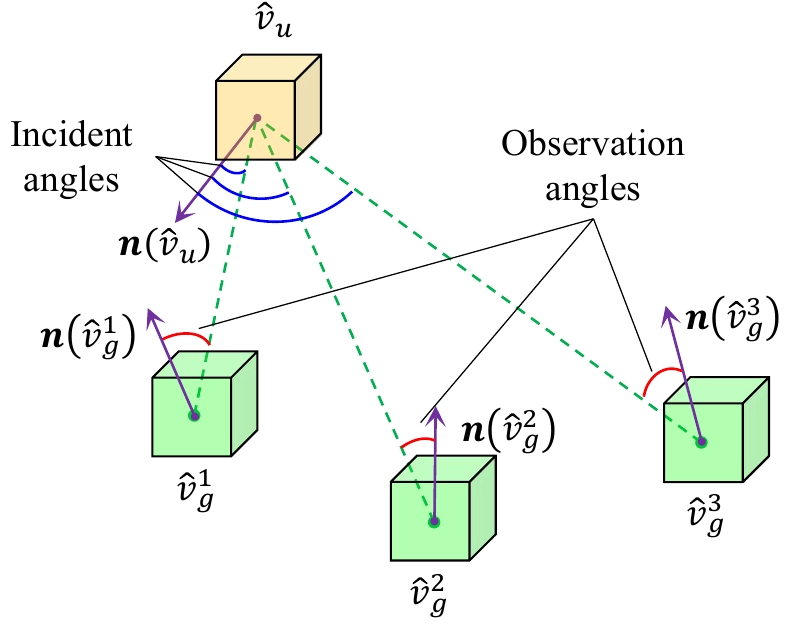}
\caption{
	Visibility check between ground and facade supervoxels. 
}
\label{fig:sv-visibility}
\end{figure}
Task views are generated with different strategies for ground and facade-ISE supervoxels. For a facade-ISE supervoxel $\hat{\bm{v}}_{u}\in\widehat{\bm{U}}_{I}$, its neighboring traversable ground-supervoxel set $\widehat{\bm{G}}^{\mathcal{N}}$ is retrieved \textcolor{black}{using a neighborhood radius defined as the maximum observation distance among all robots.} 
Each candidate view $\bm{\bm{\xi}}_{u}$ is then generated with  $\hat{\bm{v}}_{u}$ and each $\hat{\bm{v}}_{g}\in\widehat{\bm{G}}^{\mathcal{N}}$. 
A candidate view $\bm{\bm{\xi}}_{u}$ is represented by its position  $\bm{p}(\bm{\xi}_u)$, normal vector $\bm{n}(\bm{\xi}_u)$, view direction $\bm{m}(\bm{\xi}_u,\hat{\bm{v}}_u)$, traversal requirement $\bar{\zeta}(\bm{\xi}_{u})$,  and observation requirement $\bar{\varsigma}(\bm{\xi}_{u})$ (defined later in formula (\ref{eq:observationRequirement})), where 
$\bm{p}(\bm{\xi}_{u})=\bm{p}(\hat{\bm{v}}_{g})$, 
$\bm{n}(\bm{\xi}_{u})=\bm{n}(\hat{\bm{v}}_{g})$, 
$\bm{m}(\bm{\xi}_u,\hat{\bm{v}}_u)=\overrightarrow{\bm{p}(\hat{\bm{v}}_{g})\bm{p}(\hat{\bm{v}}_{u})}$, 
$\bar{\zeta}(\bm{\xi}_{u})=\bar{\zeta}(\hat{\bm{v}}_g)$. 
The normal vector $\bm{n}(\bm{\xi}_u)$ defines the 
$z$-axis orientation of the robot coordinate frame. 
Subsequently, a function is formulated to describe if the view $\bm{\xi}_u$ can observe the supervoxel $\hat{\bm{v}}_{u}$: 

\begin{equation}
\begin{split}
	\mathbb{J}(\bm{\xi}_u,\hat{\bm{v}}_u) = 
	& \text{Ray}(\bm{\xi}_u,\hat{\bm{v}}_u)
	\wedge \bigl(\text{Inc}(\bm{\xi}_u,\hat{\bm{v}}_u) < \gamma_{th}\bigr) \\
	& \wedge \bigl(\text{Obs}(\bm{\xi}_u,\hat{\bm{v}}_u) \in [\theta_{lth}, \theta_{uth}]\bigr)
\end{split}, 
\end{equation}
where the ray-casting function $\text{Ray}(\bm{\xi}_u,\hat{\bm{v}}_u)$ checks for occlusion along the line of sight in the supervoxel map, returning one if the ray is unobstructed and zero otherwise. 
The incidence-angle function $\text{Inc}(\bm{\xi}_u,\hat{\bm{v}}_u)$ computes the angle between the view direction $\bm{m}(\bm{\xi}_u,\hat{\bm{v}}_u)$ and supervoxel normal $\bm{n}(\hat{\bm{v}}_u)$. A threshold $\gamma_{th}$  is applied to reject large incidence angles that can introduce measurement errors. 
The observation-angle function $\text{Obs}(\bm{\xi}_u,\hat{\bm{v}}_u)$ measures if $\hat{\bm{v}}_u$ lies within the sensor field of view (FoV) $\bm{\xi}_u$, i.e., 
the angle between the view direction $\bm{m}(\bm{\xi}_u,\hat{\bm{v}}_u)$ and the surface normal $\bm{n}(\bm{\xi}_u)$ falls within the bounds $[\theta_{lth}, \theta_{uth}]$. The visibility check is illustrated in Fig. \ref{fig:sv-visibility}, the orange box represents the facade supervoxel, the green boxes represent the ground supervoxels, the purple arrows indicate the surface normals, the green dashed lines denote the lines of sight, the blue arcs represent the incidence angles, and the red arcs represent the observation angles. 

Subsequently, the visibility-scoring function is defined as: 
\begin{equation}
\text{Vis}(\bm{\xi}_u,\hat{\bm{v}}_u) = \frac{\mathbb{J}(\bm{\xi}_u,\hat{\bm{v}}_u)}{\beta_i \cdot \text{Inc}(\bm{\xi}_u,\hat{\bm{v}}_u) + \beta_l \cdot |\boldsymbol{m}(\bm{\xi}_u, \hat{\bm{v}}_u)|},
\end{equation}
where the incidence-angle weight is $\beta_i$, and the observation-distance weight is  $\beta_l$. Among these candidate views, the one with the highest visibility score is selected as the task view $\bm{\xi}_u^*$ of the facade-ISE supervoxel $\hat{\bm{v}}_u$, and the associated ground supervoxel of the view is $\hat{\bm{v}}_{g}^{*}$. 

After obtaining the task view of a facade-ISE supervoxel, the observation requirement $\bar{\varsigma}(\bm{\xi}_u^*)$ of the view $\bm{\xi}_u^*$ needs to be determined. 
Specifically, the robot species $k\in \bm{K}$ are sorted in ascending order of their maximum observation distance $\varsigma_{k}$ and assigned an observation-capability level $a_{\varsigma}(k)\in\{1,2,...,N_k\}$ accordingly, where the level 1 denotes the lowest requirement that all robots can satisfy. 
The observation requirement $\bar{\varsigma}(\bm{\xi}_u^*)$ is defined as the lowest capability level among all robot species that can satisfy the observation constraint for the view $\bm{\xi}_u^*$: 

\begin{equation}
\label{eq:observationRequirement}
\bar{\varsigma}(\bm{\xi}_u^*)=\min\left\{a_\varsigma(k),k\in \bm{K}\mid\left|\overline{\boldsymbol{p}(\hat{\bm{v}}_g^*)\boldsymbol{p}(\hat{\bm{v}}_u)}\right|<\varsigma_k\right\}. 
\end{equation}

Note that views generated for facade-ISE supervoxels may also cover ground-ISE supervoxels, yielding composite views that enable simultaneous observation of both supervoxel types. 
For the remaining uncovered ground-ISE supervoxels $\hat{\bm{v}}_{g}$, task views $\bm{\xi}_{g}$ are further generated. 
Assuming that a supervoxel $\hat{\bm{v}}_{g}$ is observed once a robot reaches its center $\bm{p}(\hat{\bm{v}}_g)$,  the task view $\bm{\xi}_{g}$ is defined by position $\bm{p}(\bm{\xi}_{g})$, traversal requirement  $\bar{\zeta}(\bm{\xi}_g)$, and observation requirement $\bar{\varsigma}(\bm{\xi}_{g})$, where $\bm{p}(\bm{\xi}_g)=\bm{p}(\hat{\bm{v}}_g)$, $ \bar{\zeta}(\bm{\xi}_g)=\bar{\zeta}(\hat{\bm{v}}_g)$, and $\bar{\varsigma}(\bm{\xi}_{g})=1$. 

The above process yields the facade-view set $\boldsymbol{\Xi}_{u}$ associated with the facade-ISE supervoxels ${\bm{U}}_{I}$, the ground-view set $\boldsymbol{\Xi}_{g}$ associated with the ground-ISE supervoxels ${\bm{G}}_{I}$, and the overall task-view set $\boldsymbol{\Xi}=\boldsymbol{\Xi}_{g}\cup\boldsymbol{\Xi}_{u}$. 

\subsection{Task-View Clustering} \label{sec:view_clustering}
Since the ISE supervoxels can generate numerous task views, directly solving the task-assignment problem for all views can be computationally expensive. 
Here, the task views are further grouped into task clusters for computational simplicity. 
This also allows the subsequent exploration routes to traverse regions with high view density, ensuring exploration continuity and efficiency.
However, the task views are irregularly distributed, and grouping adjacent views with different capability requirements into the same cluster can complicate subsequent task assignment. To address this issue, a task-label set $\boldsymbol{C}=\{c_0,c_1,...,c_{N_k}\}$ is defined prior to clustering. Each task view $\bm{\xi}\in \boldsymbol{\Xi}$ is classified either as a public task with label $c_{0}$ or as a private task with label $c_{k}$, where $k\in \bm{K}$ is the robot species. 
Since robots with higher capabilities can undertake more tasks and cause overload, we further introduce a requirement–capability matching function and a robot-capability scoring function. 
First, an indicator function $\mathbb{I}(\bm{\xi},k)$ is defined to represent whether the requirements of task view $\bm{\xi}$ match the capabilities of robot species $k$: 

\begin{equation} \label{eq-matchFunc}
\mathbb{I}(\bm{\xi},k)=\left(\bar{\varsigma}(\bm{\xi})<a_\varsigma(k)\right)\wedge\left(\bar{\zeta}(\bm{\xi})<a_\zeta(k)\right). 
\end{equation}
Next, the robot-capability score of each robot species $k$ is defined as:
\begin{equation} 
a(k)=\beta_{\zeta}\cdot a_{\zeta}(k)+\beta_{\varsigma}\cdot a_{\varsigma}(k), 
\end{equation}
where $\beta_{\zeta}$ and $\beta_{\varsigma}$ denote the weighting coefficients for the traversal-capability level  $a_{\zeta}(k)$ and observation-capability level $a_{\varsigma}(k)$, respectively. Then, the task views are classified as follows: 

\begin{equation} \label{eq4-11}
c(\bm{\xi})=
\begin{cases}
	c_0,      &  \text{If~} \forall k\in \bm{K},\mathbb{I}(\bm{\xi},k)=1 \\
	c_{k^*},  & \text{If~} \exists k\in \bm{K},\mathbb{I}(\bm{\xi},k)=1 \wedge k^*=\mathop{\mathrm{arg\min}}\limits_{ \substack{k\in \bm{K} \\  \mathbb{I}(\bm{\xi},k)=1} } a(k)
\end{cases}
\end{equation}
If all robot species meet the requirements of task view $\bm{\xi}$, it is labeled as a public task $c_{0}$.  
If multiple robot species satisfy the requirements, it is labeled as a private task $c_{k^*}$ of the robot species $k^*$ with the lowest robot-capability score to prevent overloading high-capability robots. 

After the task classification, clustering is performed separately for the public and private tasks defined in formula (\ref{eq4-11}). 
To prevent clustering inconsistencies among robots due to differences in supervoxel ordering, a greedy sequential clustering method is further proposed. 
First, a list is created and arranged with the hash keys of supervoxels associated with the task views. Then, the list is traversed sequentially, and elements within the clustering radius $\eta_r$ are iteratively appended to the current cluster and removed from the list. 
Once no more elements are found within  $\eta_r$, the current clustering is finished. 
The first remaining element in the list is then selected to create a new cluster, and neighboring elements within $\eta_r$ are appended to it. This process continues until the list becomes empty, resulting in a task-cluster set $\bm{T}_c=\left\{\tau_c^1,\tau_c^2,...,\tau_c^{N_\tau^c}\right\}$ for each task label $c\in\bm{C}$, where $N_\tau^c=|\bm{T}_c|$. 
The overall task-cluster set is then given by $\bm{T}=\cup_{c\in\bm{\bm{C}}}\bm{T}_c$. 
Each task cluster $\tau\in \bm{T}$ is associated with a specific task label, forming the mapping $f_c(\bm{\tau}){:}\bm{T}\mapsto \bm{C}$.  

Subsequently, the execution time of a task cluster and the transition time between task clusters are formulated. For the cluster $\tau\in \bm{T}$, the center position  $\bm{p}(\tau)=\sum_{\bm{\xi}\in\tau}\bm{p}(\bm{\xi})/|\bm{\tau}|$ is defined as the mean position of the task views within the cluster. 
The nearest node to the position $\bm{p}(\tau)$ on the topology graph $\bm{\mathcal{G}}$ \textcolor{black}{ constructed   in Sec. \ref{sec:topological_map}} is denoted by  $\bm{p}^{\prime}(\tau)=\text{neigh}_{\bm{\mathcal{G}}}(\bm{p}(\tau))$. Then, the execution time of the cluster $\tau\in \bm{T}$ for robot species $k\in \bm{K}$  is formulated as: 

\begin{equation}  \label{eq4-xx-delta}
\delta(\tau,k)=
\begin{cases} 
	\frac{\sum\limits_{\substack{\bm{\xi}\in\tau} } \text{dist}_{\bm{\mathcal{G}}}(\boldsymbol{p}(\bm{\xi}),\boldsymbol{p}^{\prime}(\tau),k)}{\vartheta_{\zeta}(k)}+\frac{|\boldsymbol{\Xi}_{u}^{\tau}|}{\vartheta_{\varsigma}(k)}, & \text{If~} f_{c}(\tau)\in\{c_{0},c_{k}\} \\
	+\infty, & \text{Otherwise}
\end{cases}, 
\end{equation}
where $\vartheta_\zeta(k)$ is the average movement velocity of robot species $k$, and $\vartheta_{\varsigma}(k)$ is the average observation velocity measured as the mean number of facade task views covered per unit time.  
The subset of task views that belong to both the cluster $\tau$ and facade-view set $\boldsymbol{\Xi}_{u}$ is denoted as $\boldsymbol{\Xi}_{u}^{\tau}=\{\bm{\xi}|\bm{\xi}\in\tau\wedge\boldsymbol{\bm{\xi}}\in\boldsymbol{\Xi}_{u}\}$. The distance query function $\text{dist}_{\bm{\mathcal{G}}}(\cdot,\cdot,\cdot)$ is defined in Sec. \ref{sec:topological_map}. Furthermore, the transition time from task cluster $\tau_{i}\in \bm{T}$ to task cluster $\tau_{j}\in \bm{T}$ for robot species $k\in\bm{K}$ is defined as:

\begin{equation} \label{eq4-xx-w}
w(\tau_i,\tau_j,k)=\text{dist}_{\bm{\mathcal{G}}}(\boldsymbol{p}^{\prime}(\tau_i),\boldsymbol{p}^{\prime}(\tau_j),k)/\vartheta_\zeta(k).
\end{equation}


\subsection{Heterogeneous Multi-Robot Collaborative Task Assignment} \label{sec:hdmtsp}

\subsubsection{Problem Formulation}\label{sec:problemDefinition}
The assignment of task clusters to robots is formulated as a HMDMTSP \cite{miloradovic2021gmp}, where the terminology is redefined by replacing salesmen with robots, cities with task clusters, and depots with robot starting positions. The route of each robot starts from its current position and is not required to return to the starting point, resulting in an open-path formulation. 
Herein, we further incorporate the matching constraints between robot-species capabilities and task-cluster requirements. 
Recall that the robot team, denoted as the set $\bm{R}=\{r_1,r_2,...,r_{N_r}\}$ , consists of $N_r$ robots $r\in\bm{R}$. 
The task-cluster set $\boldsymbol{T}=\{\tau_{1},\tau_{2},...,\tau_{N_{\tau}}\}$ consists of  $N_\tau$ clusters $\tau \in \bm{T}$. 
We further introduce the set of robot-starting positions $\boldsymbol{\Sigma}=\{\sigma_{1},\sigma_{2},...,\sigma_{N_{r}}\}$, containing $N_{r}$ positions  $\sigma\in\boldsymbol{\Sigma}$.
A new graph is formulated as  $\widetilde{\bm{\mathcal{G}}}=(\widetilde{\bm{\mathcal{V}}},\widetilde{\bm{\mathcal{E}}})$, where $\widetilde{\boldsymbol{\mathcal{V}}}=\boldsymbol{T}\cup\boldsymbol{\Sigma}$ denotes the union of all task clusters and robot-starting positions, with a total of $N_{\tilde{\mathcal{V}}}$ nodes.  
The edge set $\widetilde{\bm{\mathcal{E}}}$ is derived from the topology graph $\bm{\mathcal{G}}$ \textcolor{black}{ constructed   in Sec. \ref{sec:topological_map}}. 
The problem aims to find Hamiltonian routes for the robot team such that all vertices in $\widetilde{\bm{\mathcal{G}}}$ are visited exactly once by type-compatible robots while minimizing the objective function. 
The optimization is typically cast as a mixed-integer linear program, which is known to be NP-hard, and the inclusion of additional requirement-capability constraints substantially increases computational complexity.

Inspired by \cite{miloradovic2021gmp}, we propose an improved genetic algorithm to efficiently solve the problem. Since the original genetic algorithm \cite{miloradovic2021gmp} iteratively refines candidate solutions through operations such as mutation, selection, and crossover, its effectiveness depends on both the solution representation and fitness function. 
Based on the multi-chromosome representation \cite{kiraly2011optimization}, we further design population generation, individual mutation operations, and a fitness function for heterogeneous multi-robot task assignment. 
\textcolor{black}{
Since decentralized collaborative exploration planning requires each robot to generate routes while considering the team's task assignments, additional constraints are incorporated to preserve global consistency. 
} 

\subsubsection{Population Generation}\label{sec:populationGeneration}
\begin{figure}[tpb]
\centering
\includegraphics[width=1\linewidth]{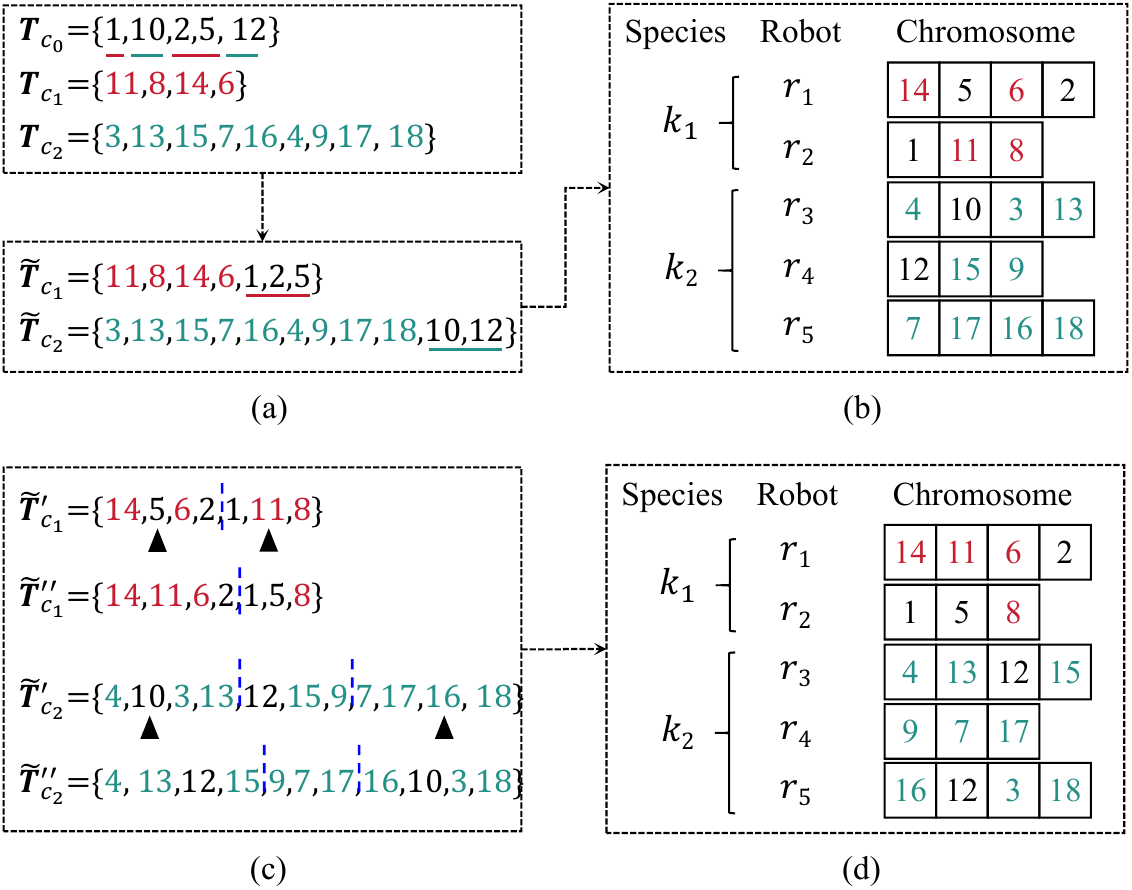}
\caption{
	Stochastic individual generation of multi-chromosome and mutation operations. 
	(a) Randomly separate the public task clusters into private task clusters. 
	(b) Generate individuals randomly from the private task clusters. 
	(c) Regenerate the private task clusters from the individuals in (b), followed by swap and sliding operations. 
	(d) New individuals corresponding to the mutated private task clusters in (c).
}
\label{fig:mutation}
\end{figure}

An individual, i.e., a potential solution to the problem, is represented by a task-cluster sequence. 
At the beginning of an iteration, the initial population is obtained by randomly creating many individuals as follows. 
First, the elements of the public task-cluster set $\bm{T}_{c_0}$ are randomly assigned to the private task-cluster sets $\{\bm{T}_{c_k}, k\in\bm{K}\}$, resulting in augmented private task-cluster sets $\{\widetilde{\bm{T}}_{c_k}, k\in\bm{K}\}$. 
Then, the views in each augmented private task-cluster set are randomly split and assigned to multiple robots with proper capabilities, thereby forming an assignment scheme, i.e., an individual.  
By doing this, the assignment of public task clusters to the augmented private task-cluster sets is determined during the initial population generation. 
A sufficiently large population size is adopted to ensure the diversified assignment of public task clusters. 
Figs. \ref{fig:mutation}(a) and (b) illustrate the generation of an individual corresponding to three task clusters and five robots of two species. As shown in Fig. \ref{fig:mutation}(a), the set of public task clusters $\bm{T}_{c_0}$  is split and assigned to the private task-cluster sets $\bm{T}_{c_1}$ and $\bm{T}_{c_2}$, resulting in the augmented ones $\widetilde{\bm{T}}_{c_1}$ and $\widetilde{\bm{T}}_{c_2}$. Fig. \ref{fig:mutation}(b) shows the individual, i.e., multiple chromosomes, obtained after randomly assigning the public task-cluster sets  $\widetilde{\bm{T}}_{c_1}$ and $\widetilde{\bm{T}}_{c_2}$.

\subsubsection{Mutation Operations}\label{sec:mutationOperations}
The original genetic algorithm \cite{miloradovic2021gmp} employs mutation and crossover operations to enhance population diversity. 
For small-scale task-assignment problems, relying solely on mutation and selection operations is sufficient to thoroughly explore the solution space \cite{kiraly2011optimization}. 
Typical mutation operations include swap, shift, inversion, shuffle, and combinations of shuffle with other ones. 
In our multi-chromosome representation, the mutation process is redesigned. 
Since the assignment of public task-cluster sets is determined once initial populations are obtained, the mutations further focus on the assignment of augmented private task-cluster sets. 
Specifically, for an individual, the task clusters assigned to robots of the same species are first merged to reconstruct augmented private task-cluster sets. Mutation operators, including swap, shift, and shuffle, are then applied to each task-cluster set. 
Finally, the mutated task-cluster sets are  redistributed among the robots to form new individuals. 
\textcolor{black}{Figs. \ref{fig:mutation}(b) and (c)} illustrate an example of mutation, where the assignments of robots with the same species are merged to form new augmented private task-cluster sets, denoted as $\widetilde{\bm{T}}_{c_1}^{\prime}$ and $\widetilde{\bm{T}}_{c_2}^{\prime}$. As illustrated in Fig. \ref{fig:mutation}(c)), 
a swap operation is performed on the task-cluster set  $\widetilde{\bm{T}}_{c_1}^{\prime}$, where the triangles indicate the two randomly selected elements to be exchanged. Similarly, a shift operation is applied to the task-cluster set $\widetilde{\bm{T}}_{c_2}^{\prime}$, in which the two triangles indicate the randomly selected interval. In this example, the elements within the interval are cyclically shifted left by two positions. Fig. \ref{fig:mutation}(d) demonstrates the regenerated individual based on the mutated private task-cluster sets $\widetilde{\bm{T}}_{c_1}^{\prime\prime}$ and $\widetilde{\bm{T}}_{c_2}^{\prime\prime}$. 
The proposed individual definition and mutation operations reduce the search space and accelerate problem-solving. 


\subsubsection{Fitness Function}\label{sec:fitnessFunction}
The quality of an individual that represents a feasible solution to the problem is evaluated by the fitness function. 
An individual  $\boldsymbol{h}=\left\{\widetilde{\bm{\mathcal{V}}}_{r_{1}},\widetilde{\boldsymbol{\mathcal{V}}}_{r_{2}},...,\widetilde{\boldsymbol{\mathcal{V}}}_{r_{N_{r}}}\right\}$ consists of the task-cluster sequences assigned to each robot $r\in\bm{R}$, where the sequence of robot $r$  is defined as  $\widetilde{\boldsymbol{\mathcal{V}}}_r=\left\{\sigma_r,\tau_r^1,\tau_r^2,...,\tau_r^{N_r^\tau}\right\}\subseteq\widetilde{\boldsymbol{\mathcal{V}}}$ with  $|\widetilde{\boldsymbol{\mathcal{V}}}_r|=N_{r}^{\tau}+1$. The entire set of assigned clusters is $\widetilde{\boldsymbol{\mathcal{V}}}=\cup_{r\in R}\widetilde{\boldsymbol{\mathcal{V}}}_r$. 
Inspired by \cite{soar}, we incorporate a total execution cost, a maximum execution cost, and a distance to reference routes into the fitness function to prioritize superior individuals. 

First, the execution cost of a task-cluster sequence $\widetilde{\boldsymbol{\mathcal{V}}}_r$ is defined as the sum of the transition time between consecutive task clusters and the execution time of each task cluster: 
\begin{align}	\label{eq:cost_function}
\text{cost}(\widetilde{\boldsymbol{\mathcal{V}}}_r) 
&= w(\sigma_r, \tau_r^1, f_k(r)) 
+ \sum_{i=2}^{N_r^\tau} w(\tau_r^{i-1}, \tau_r^i, f_k(r)) \notag \\
&\quad + \sum_{i=1}^{N_r^\tau} \delta(\tau_r^i, f_k(r)), 
\end{align}
where $\delta(\cdot,\cdot)$  denotes the task-cluster execution time defined in formula (\ref{eq4-xx-delta}) (see Sec. \ref{sec:view_clustering}), $w(\cdot,\cdot,\cdot)$ represents the transition time between two task clusters defined in formula (\ref{eq4-xx-w}) (see Sec. \ref{sec:view_clustering}), 
\textcolor{black}{and $f_{k}(r)$ maps the robot $r\in\bm{R}$ with an associated species $k\in\bm{K}$ (see Sec. \ref{sec:map_representation_fusion})}. 

\textcolor{black} {Aiming at constructing a decentralized collaborative framework, task assignment must be independently computed by each robot based on global information. }
However, the randomness of the mutation operation leads to inconsistent solutions across multiple robots, resulting in route overlap or area abandonment and degrading exploration efficiency. To ensure solution consistency across robots, 
\textcolor{black}{the solution $\check{\bm{h}}_{r}$ obtained by each robot $r\in\bm{R}$ in the previous iteration is shared}. The solutions are then merged into a set  $\check{\bm{\mathcal{H}}}=\begin{Bmatrix}\check{\boldsymbol{h}}_{r_1},\check{\boldsymbol{h}}_{r_2},...,\check{\boldsymbol{h}}_{r_{N_r}}\end{Bmatrix}$, from which the optimal solution $\check{\boldsymbol{h}}^{*}$  is selected according to a unified rule: 
\begin{equation}
\label{eq:minMax}
\check{\boldsymbol{h}}^*=\arg\min_{\check{\boldsymbol{h}}_r\in\check{\boldsymbol{H}}}\left(\max_{\check{\boldsymbol{\widetilde{\mathcal{V}}}}_i\in\check{\boldsymbol{h}}_r}\text{length}\left(\check{\widetilde{\boldsymbol{\mathcal{V}}}}_i\right)\right), 
\end{equation}
where the function $\text{length}(\cdot)$ calculates the route length of a task-cluster sequence. 
This rule prevents the selection of excessively long routes, thereby avoiding detours and ensuring exploration efficiency. 
The optimal solution $\check{\boldsymbol{h}}^*$ serves as the reference route for the current planning round, ensuring consistency among robots in the selection of reference routes. 

Subsequently, a distance function is defined between each robot’s task-cluster sequence \textcolor{black}{$\widetilde{\bm{\mathcal{V}}}_r$} in the current solution  $\bm{h}$ and the corresponding reference sequence  $\check{{\boldsymbol{\mathcal{\widetilde{\mathcal{V}}}}}}^*_r$ in $\check{\boldsymbol{h}}^*$: 


\begin{equation}
\label{eq:fitness}
\text{dist}\left(\widetilde{\boldsymbol{\mathcal{V}}}_r,\check{\widetilde{\boldsymbol{\mathcal{V}}}}_r^*\right) = \sum_{i=1}^{|\widetilde{\boldsymbol{\mathcal{V}}}_r|} \left(e^{-\beta_e\cdot i}\cdot\min_{\tau_j,\tau_{j+1}\in\check{\widetilde{\boldsymbol{\mathcal{V}}}}_r^*}\mathcal{L}(\tau_i,\tau_j,\tau_{j+1}) \right), 
\end{equation}
where the exponential weighting factor $\beta_{e}$ assigns greater importance to task clusters located earlier in the sequence. The function $\mathcal{L}(\cdot,\cdot,\cdot)$ computes the minimum Euclidean distance from a point to a line segment. 
Specifically, consecutive task clusters in the reference route form a line segment $\overline{\bm{p}(\tau_j)\bm{p}(\tau_{j+1})}$, and the minimum distance from each task-cluster center $\bm{p}(\tau_i)$ in the current solution to all such segments of the reference route is then calculated. 

Finally, the total execution cost, the maximum execution cost, and the penalty term with respect to the reference routes are incorporated into the fitness function: 
\begin{align}
\bm{\mathcal{J}}(h)
&=\beta_c \cdot \sum_{ r\in\bm{R} } \text{cost} (\widetilde{\boldsymbol{\mathcal{V}}}_r) + \beta_m \cdot \max_{ r\in\bm{R} } \text{cost}(\widetilde{\boldsymbol{\mathcal{V}}}_r) \notag \\
&\quad + \beta_d \cdot \sum_{ r\in\bm{R} } \text{dist} \left( \widetilde{\boldsymbol{\mathcal{V}}}_r,\check{\widetilde{\boldsymbol{\mathcal{V}}}}_r^* \right), 
\end{align}
where $\beta_{c}$ weights the total execution time,  $\beta_m$  weights the maximum execution time, and  $\beta_d$ weights the reference-route distance. The first term of the fitness function ensures the compactness of the robot routes, the second term promotes a more balanced workload among robots, and the third term encourages the newly generated solutions to follow the reference routes, thereby alleviating the influence of route randomness. 

\subsection{Exploration Tour Generation} \label{sec:exploration_tour_generation}
The task-cluster assignment is further utilized to generate exploration routes for robots. 
Since co-visibility relationships among views are not considered during view generation, the views within each cluster can be further refined to enable more efficient traversal.
Herein, we adopt the MNSS \cite{mytim} method to simplify each task cluster $\tau_r^i\in\widetilde{\boldsymbol{\mathcal{V}}}_r$. 
First, task views are generated for the traversable ground supervoxels within the neighborhood $\eta_r$ of each task-cluster center and are added into the corresponding cluster. Then, based on the visibility between the views within the cluster and their associated ISE supervoxels, a score matrix is constructed. Views are sampled with probabilities proportional to their scores to obtain a view set that fully covers the ISE supervoxels, and a view sequence with minimal execution cost is produced, as defined in formula (\ref{eq:cost_function}). Through the MNSS, adjacent views are rejected to reduce redundant observations while maintaining both the traversal and observation constraints of robot species. 
To ensure that the final view sequence preserves the execution order of the task clusters, the center point of the previous task cluster  $\bm{p}(\tau_r^{i-1})$ is set as a virtual start point, and the center point of the next task cluster $\bm{p}(\tau_r^{i+1})$ is set as a virtual end point. As a result, the MNSS regenerates the view sequence within each cluster:  $\tilde{\tau}_{r}^{i}=\left\{\tilde{\bm{\xi}}_{r}^{i,1},\tilde{\bm{\xi}}_{r}^{i,2},...,\tilde{\bm{\xi}}_{r}^{i,\left|\tilde{\tau}_{r}^{i}\right|}\right\}$. 
By generating view sequences for all task clusters, a complete task-view sequence, i.e., an exploration route, is finally obtained: 
\begin{align}
\boldsymbol{\Xi}_r 
& = \{ \tilde{\boldsymbol{\xi}}_r^{1,1},\tilde{\boldsymbol{\xi}}_r^{1,2},...,\tilde{\boldsymbol{\xi}}_r^{1,\left|\tilde{\boldsymbol{\tau}}_r^1\right|},\tilde{\boldsymbol{\xi}}_r^{2,1},\tilde{\boldsymbol{\xi}}_r^{2,2},...,\tilde{\boldsymbol{\xi}}_r^{2,\left|\tilde{\boldsymbol{\tau}}_r^2\right|},..., \notag \\
& \quad\quad
\tilde{\boldsymbol{\xi}}_r^{N_r^\tau,1},\tilde{\boldsymbol{\xi}}_r^{N_r^\tau,2},...,\tilde{\boldsymbol{\xi}}_r^{N_r^\tau,\left|\tilde{\boldsymbol{\tau}}_r^{N_r^\tau}\right|} \}.
\end{align}

\begin{figure}[t]
\centering
\includegraphics[width=0.98\linewidth]{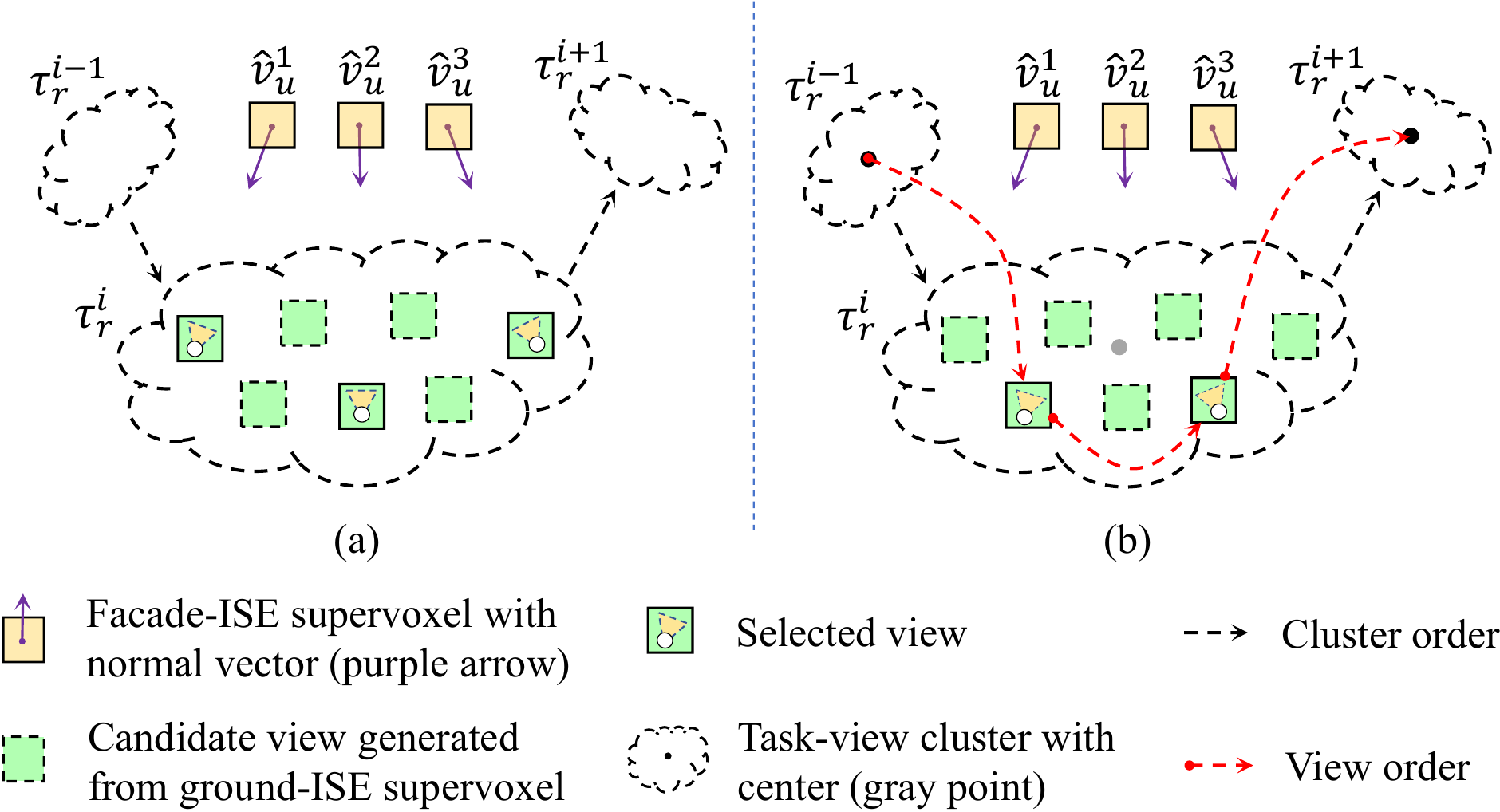}
\caption{
    \textcolor{black} {  
	Generation process from task-view clusters to task-view sequences.   
    (a) Candidate views are produced by extracting ground-ISE supervoxels near the current cluster. 
    (b) A new view sequence is obtained via the MNSS.     }
}
\label{Fig.4.7.ExplorationPlanning}
\end{figure}

\textcolor{black} {Figure \ref{Fig.4.7.ExplorationPlanning} } illustrates the generation process from task-view clusters to task-view sequences. As illustrated in Fig. \ref{Fig.4.7.ExplorationPlanning}(a),  the task views (green solid boxes) within the current cluster $\tau_r^i$ are associated with the facade-ISE supervoxels (orange boxes), and candidate views (green dashed boxes) are generated with the traversable ground supervoxels in the neighborhood  \textcolor{black} {(see Sec. \ref{sec:task_view_generation}). }
As shown in Fig. \ref{Fig.4.7.ExplorationPlanning}(b), candidate views are sampled to regenerate task views (green solid boxes) \textcolor{black} {using the MNSS}. 
With the center of previous task cluster $\bm{p}(\tau_r^{i-1})$ set as the virtual start point and the center of next task cluster $\bm{p}(\tau_r^{i+1})$ set as the virtual end point, the traversal sequence (red dashed arrows) of task views within the current cluster is obtained. 

\subsection{Path-Conflict Resolution Strategy} \label{sec:resolve_path_conflicts}
Based on the task-view sequences assigned to each robot, executable motion paths are further planned to approach task views and acquire measurements sequentially. 
However, relying solely on individual motion planning cannot guarantee safety in multi-robot systems because of path intersections.
To address this issue, the robots are required to exchange their executable paths to resolve conflicts. Since collisions occur only when robots are in proximity, broadcasting only the portion of their paths within a distance $\eta_p$ of their current positions is sufficient to detect conflicts while reducing communication bandwidth. 
Herein, path conflicts are resolved with two strategies: 
\subsubsection{Collision Avoidance}
When multiple robots approach each other along intersecting motion paths, the time-to-collision is estimated based on their velocities and distances to the crossing point. The robot predicted to arrive later should stop and maintain a safe distance, allowing the earlier-arriving robot to pass first. 
\subsubsection{Obstacle Avoidance}
When a robot stops for collision avoidance or region scanning, other robots treat it as an obstacle and replan motion paths to bypass it. This is achieved by creating a bounding box around the stationary robot and penalizing waypoints close to it during motion planning. 

\section{Simulations and Experiments}  
The proposed approach was comprehensively evaluated and compared with state-of-the-art approaches, and ablation studies were conducted to validate its components. Simulations analyzed performance across diverse scenes and heterogeneous robot teams, while the real-world experiments validated the feasibility and robustness of the proposed approach.   

\subsection{Multi-Robot Configurations}
\label{sec:experiments}\begin{figure}[t]
\centering
\includegraphics[width=0.9\linewidth]{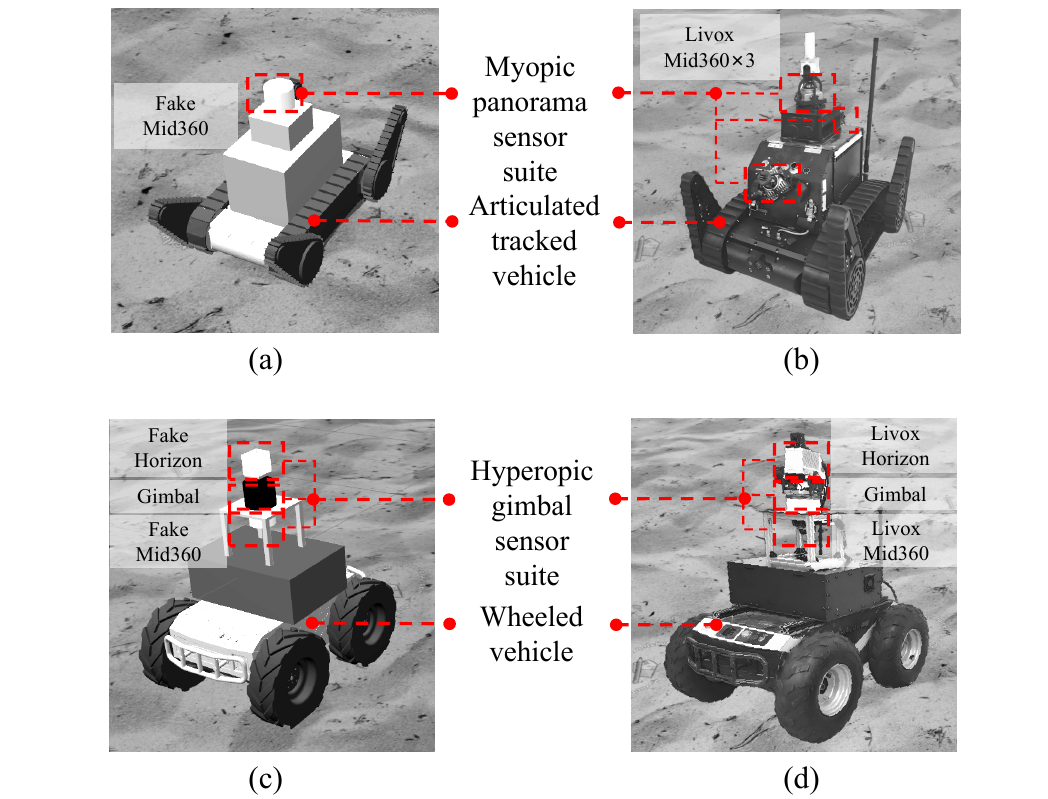}
\caption{
	\textcolor{black}{Robots} in simulation and the real world. (a)(b) The articulated tracked robot in simulation and the real world. (c)(d) The wheeled robot in simulation.
}
\label{fig:robots}
\end{figure}
Two typical robot species with different mobility and observation capabilities are employed to form a heterogeneous multi-robot team for evaluations. As shown in Figs. \ref{fig:robots}(b) and (d), the two robot platforms consist of an articulated tracked robot and a wheeled robot, equipped with custom-designed sensor suites to improve perception abilities. The articulated tracked robot offers superior traversal capability and is well-suited for rugged terrain in both indoor and outdoor environments, whereas the wheeled robot achieves higher movement speed but is limited to flat terrain. 
The articulated tracked robot is equipped with a myopic panoramic sensor suite comprising three Livox Mid360 LiDARs mounted at specific positions. With the non-repetitive scanning pattern and the stitched fields of view of the Livox Mid360 (vertical 60$^\circ$, horizontal 360$^\circ$), the sensor suite achieves panoramic dense scanning (see Fig. \ref{fig:robots}(b)). However, due to the limited sensing range of the Livox Mid360 (up to 40 m), this suite is suitable only for indoor or small-scale outdoor scenes.  
The wheeled robot carries a hyperopic gimbal-sensor suite, comprising a Livox Horizon and a Livox MID360 LiDARs as well as a gimbal (see Fig. \ref{fig:robots}(d)). Since the Livox Horion has a long sensing range (up to 90 m) but a narrow FoV (vertical ±11.25°, horizontal ±45°), a gimbal with yaw and pitch degrees of freedom is added to increase observation flexibility, allowing the wheeled robot to effectively scan the high-rise walls of buildings. The Livox Mid360 is installed upside-down at the bottom of gimbal, which efficiently scans near-ground areas and mitigates localization degradation caused by the limited FoV of Livox Horizon.

In simulations, the Gazebo\footnote{https://gazebosim.org/} simulator was adopted to construct the robot team and urban building environments. Due to the high computational overhead of the official Livox simulation plugins\footnote{https://github.com/Livox-SDK/livox\_laser\_simulation}, several changes were made to improve real-time performance. 
For the myopic panoramic sensor suite, three Livox Mid360 plugins were replaced with a 120-line rotating LiDAR plugin with a horizontal 360$^\circ$ and vertical ±45$^\circ$ FoV to achieve comparable point-cloud density and scanning coverage. 
For the hyperopic gimbal-sensor suite, a 32-line rotating LiDAR plugin with horizontal and vertical FoVs of ±45$^\circ$ and ±11.25$^\circ$, respectively, was used to emulate the Livox Horizon. In contrast, an 80-line rotating LiDAR plugin with a horizontal FoV of -45$^\circ$ to +15$^\circ$ was employed to replace the Livox Mid360 plugin.

\begin{table*}[b]
\caption{Parameters of the proposed approach}
\label{tab:parameters2}
\centering
\begin{tabular}{|c|c|c|c|c|}
	\hline
	\textbf{Parameter} & \textbf{Value} \\ \hline
	
	Robot species & $k_1$=SLUGCAT, $k_2$=MR1000 \\  \hline
    Observation-density threshold  & $\rho_{th}$=0.3 \\ \hline
    Observation-angle threshold  & $\gamma_{th}$=45° \\  \hline
    Observation-distance threshold  & $l_{th}$=10m (SLUGCAT), 27m (MR1000) \\  \hline
    Ground-facade voxel slope threshold  &  $s_{th}$=45° \\ \hline
    Bounds of the observation-angle function  & $\theta_{lth}$=$-45^\circ$, $\theta_{uth}$= $+45^\circ$ \\  \hline
    
	Maximum observation distance & $\varsigma_{k_1}$=10 m, $\varsigma_{k_2}$=27 m \\  \hline
	Average mobility velocity & $\vartheta_{\zeta}(k_{1})$=0.4 m/s, $\vartheta_{\zeta}(k_{2})$=0.5 m/s \\  \hline
	Average observation velocity & $\vartheta_{\varsigma}(k_{1})$=10 pieces/s, $\vartheta_{\varsigma}(k_{2})$=0.1 pieces/s \\  \hline

	Incidence-angle weight in visibility-scoring function  & $\beta_i$=0.5 \\  \hline
	Observation-distance weight in visibility-scoring function   & $\beta_l$=0.5 \\  \hline
	Traversal-capability-level weight in robot-capability score  & $\beta_\zeta$=10 \\  \hline
	Observation-capability-level weight in robot-capability score  & $\beta_\varsigma$=1 \\  \hline

	Exponential distance weight for reference route and task clusters  & $\beta_e$=0.5 \\  \hline
	Total execution-time weight  in the fitness function & $\beta_c$=1.0 \\  \hline
	Maximum execution-time weight in the fitness function  & $\beta_m$=1.5 \\ \hline
	Reference-route distance weight in the fitness function & $\beta_d$=3 \\  \hline
	
	Supervoxel-seed merging distance  & $\eta_e$=0.2 m \\ \hline
	Supervoxel-seed interval  & $\eta_s$=1 m \\ \hline
	High-level map synchronizing period  & $\eta_c$=1 s \\ \hline
	Task-view clustering radius  & $\eta_r$=5 m \\ \hline	Executable path truncation length  & $\eta_p$=2 m \\ \hline
	Route-execution timeout  & $\eta_t$=10 s \\ \hline
\end{tabular}
\end{table*}

The simulations used ground-truth odometry from the Gazebo plugin, while the field experiments employed onboard sensors, with each robot independently running SLAM for pose estimation. Additionally, to achieve inter-robot relative localization in the real world, robots were initialized in proximity to each other for map registration at the beginning of exploration. 
Subsequently, each robot updated its map in the common world frame. 
Both simulations and experiments employed the same terrain-aware path-planning approach \cite{mytro}. To ensure fair evaluation and seamless transfer to real robots, the simulations and physical deployments of the proposed approach were executed on the robots' onboard computers  (Intel Core i9-10900K CPU with 32 GB RAM). The parameters are listed in Table \ref{tab:parameters2}.

\subsection{Decentralized Multi-Robot Communication} \label{sec:sec_comm_framework}
\subsubsection{Software and Hardware Environment}
The proposed multi-robot system adopted a hybrid ROS 1/ROS 2 architecture. 
Given that ROS1 provides a mature ecosystem for robotic development, the main programs of the proposed framework were deployed on ROS1. Meanwhile, ROS2 was employed as the decentralized middleware to leverage its advanced data distribution service. The ``ros1\_bridge" package was utilized to achieve seamless interoperability between ROS1 and ROS2 nodes. 
Furthermore, different quality-of-service policies were applied in ROS2 according to the topic characteristics: high-real-time topics, e.g., robot-state and sensor-data topics, were assigned low-latency policies, while control-command topics were set to high-reliability profiles, thereby satisfying the communication requirements of the multi-robot system. 
For data transmission, a wired local area network was implemented in simulations using Ethernet switches, and a wireless self-organized network was employed in real-world experiments, with each robot equipped with a radio terminal. 

\subsubsection{Decentralized Multi-Robot Simulation Environment}
Given the limited computational resources of a single computer, simultaneously loading a large-scale complex scene, spawning multiple robots, and executing mapping, planning, and exploration modules would severely overload the CPU/GPU, leading to simulation crashes or unacceptable slowdowns. 
To address this issue, a fully decentralized simulation architecture was implemented using the Gazebo simulator, distributing the computational load across multiple computers. Specifically, each computer loaded the same world file and instantiated all robot models to ensure an identical simulation environment, but performed simulations and modules only for its designated robot. Each computer continuously broadcast the states of its designated robot over the network and simultaneously received the states of remote robots from other computers. 
Lightweight kinematic proxies of the remote robots were driven directly by the received data, with no dynamics or sensor simulation. This enabled decentralized collaborative multi-robot simulations to be performed across multiple computers while achieving a high real-time factor. 

\subsubsection{Human-Robot Interaction Interface on Ground Station}
For comprehensive maintenance of the multi-robot system, a ground station was established for human-robot interaction, without interfering with the robots’ decentralized planning processes. Specifically, it was dedicated to publishing high-level mission commands, such as starting exploration tasks or triggering emergency stops, and to providing systematic visualization. 
It collected shared information from all robots—such as current pose, historical positions, planned paths, built maps, communication latency, and signal strength—and provided operators with an intuitive, real-time, and holistic monitoring interface. Furthermore, it supplied high-precision timing synchronization to all robots, thereby avoiding data misalignment caused by clock drift.  

\begin{table*}[b]
\caption{Principles of the comparative approaches and their heterogeneous adaptations}
\label{tab4_3_sota_comarision}
\centering
\renewcommand{\arraystretch}{1.2}  
\begin{tabular}{|m{1.7cm}|m{1.5cm}|m{3cm}|m{3cm}|m{3cm}|m{3cm}|}
	\hline
	\centering\textbf{Module} & \centering\textbf{Attribute} & \centering\textbf{MDVRP} \cite{yan2022mui} & \centering\textbf{MROMT} \cite{dong2019multi} & \centering\textbf{TSPGA} \cite{hardouin2023multirobot} &  \multicolumn{1}{c|}{\textbf{Ours}} \\ 		\hline
	
	\multirow{2}{*}{\begin{tabular}[c]{@{}c@{}} Map \\ Representation \end{tabular}}
	& \centering Principle & 3D volumetric map, sub-space division, pose graph & 2D ESDF map, validity map, uncertainty map & 3D TSDF map & 3D basic perception map, supervoxel map, traversal-topology graph \\
	\cline{2-6}
	& \centering heterogeneity & \centering No & \centering No & \centering No & \multicolumn{1}{c|}{\textbf{Yes}} \\
	\hline
	\multirow{2}{*}{\begin{tabular}[c]{@{}c@{}} ~~View \\ ~~~Generation \end{tabular}} & \centering Principle & Evaluate information gain of candidate views and generate the shortest coverage route via probabilistic sampling. & Construct a priority queue based on validity and uncertainty scores to select task-view sequences. & Extract ISE to guide view generation with TSDF parameters, calculate information gain to filter view clusters & Generate task views with ISE supervoxels and analyze the traversal and observation requirements of views. \\
	\cline{2-6}
	& \centering heterogeneity & \centering No & \centering No & \centering No & \multicolumn{1}{c|}{\textbf{Yes}} \\
	\hline
	\multirow{2}{*}{\begin{tabular}[c]{@{}c@{}} ~~Task \\ ~~Assignment \end{tabular}} & \centering Principle 
    & Model a MDVRP to solve sub-space assignment and traversal order for robots, with a ``min-max" strategy to improve consistency 
    & Model an optimal mass-transport problem to assign task views, minimizing total cost and balancing loads 
    & Formulate a TSP to solve the traversal route and assign task-view clusters with information gain via a greedy strategy 
    & Formulate an HMdMTSP to solve task-view cluster assignment and execute minimal view set sampling to simplify views \\
	\cline{2-6}
	& \centering heterogeneity & \centering No & \centering No & \centering No & \multicolumn{1}{c|}{\textbf{Yes}} \\
	\hline
	\centering Communication & \centering Architecture & \centering Centralized & \centering Centralized & \centering \textbf{Decentralized} & \multicolumn{1}{c|}{\textbf{Decentralized}} \\
	\hline
	\multirow{2}{*}{\begin{tabular}[c]{@{}c@{}} Heterogeneous \\ Adaptions \end{tabular}} & \centering Principle 
	& Replace the original map representation and view generation with ours, 
	assign task views rather than sub-space, and introduce task-view requirements in the cost matrix for the MDVRP solving. 
	& Replace the original map representation and view generation with ours, inherit the priority queue to select task views, and incorporate requirement-capability constraints into the cluster assignment. 
	& Replace the original map representation and view map generation with ours, inherit the TSPGA for assignment, and adopt our greedy sequential clustering to ensure consistent clustering among robots. & \multicolumn{1}{c|}{-} \\
	\cline{2-6}
	& \centering Variant Name & \centering MDVRP+ & \centering MROMT+ & \centering TSPGA+ & \multicolumn{1}{c|}{-} \\
	\hline
\end{tabular}
\end{table*}

\subsection{Comparative Approaches}
Comprehensive evaluations were conducted in simulations for the proposed approach and three state-of-the-art approaches. 
Most existing multi-robot exploration approaches are designed for homogeneous robots, whereas their map-representation and view-generation modules are not suitable for collaborative exploration in complex 3D environments. 
Herein, three representative approaches with the scalability to heterogeneous multi-robot task assignment were selected for comparison, including the multi-depot vehicle routing problem-based approach (MDVRP)\cite{yan2022mui}, the multi-robot optimal mass transport-based approach (MROMT) \cite{dong2019multi}, and the traveling salesman problem-greedy assignment-based approach (TSPGA) \cite{hardouin2023multirobot}. The principles of these approaches are analyzed in Table \ref{tab4_3_sota_comarision}. 
The MDVRP approach models the task assignment for multi-robot exploration as a mixed-integer linear program, which is solved using OR-Tools\footnote{https://developers.google.com/optimization}. It also introduces a ``min-max" strategy that selects the best solution from the previous planning round as a reference for the current round, thereby mitigating inconsistencies in multi-robot routes caused by solver randomness. 
The TSPGA approach assigns task-view clusters using a greedy strategy that prioritizes the robot with the largest information-gain increments. 
The MROMT approach constructs priority queues to select task views based on validness and uncertainty scores. It formulates task assignment as an optimal mass-transport problem, solved with a modified Lloyd algorithm. 

Since the comparative approaches were originally implemented for simplistic robot models and flat exploration regions, their mapping and planning modules are unsuitable for traversing indoor-outdoor regions with uneven terrains. 
To ensure fair comparison, several modules of the comparative approaches were replaced with our proposed ones, such as the map representation and fusion module (see Sec. \ref{sec:map_representation_fusion}), the task-view generation module (see Sec. \ref{sec:task_view_generation}), and the communication architecture (see Sec. \ref{sec:sec_comm_framework}). Moreover, the task-assignment modules of comparative approaches were marginally modified to support heterogeneous robots. 
The variant approaches are denoted as MDVRP+, TSPGA+, and MROMT+, and their modified task-assignment modules are detailed as follows. 
The MDVRP+ approach introduced constraints between task views and robot capabilities into the cost matrix to enable proper task-assignment problem solving. 
The TSPGA+ approach incorporated the view-clustering method described in Sec. \ref{sec:view_clustering} before performing task-cluster assignment to ensure clustering consistency across robots. 
The MROMT+ approach integrated co-visibility relationships into the task-view scoring and priority-queue selection to reduce redundant views, followed by a clustering-based algorithm to compute optimal traversal routes. 
With the above modifications, all comparative approaches were endowed with identical capabilities for navigating challenging terrain, and thus the comparisons were enabled to emphasize the effectiveness of task assignment in heterogeneous multi-robot exploration. Additionally, ablation studies were further conducted to comprehensively validate each component of the proposed framework (see Sec. \ref{sec:ablationStudy}), forming a systematic evaluation.


\subsection{Multi-Robot Collaborative Exploration Simulations}

\begin{figure}[]
    \centering
    \includegraphics[width=1\linewidth]{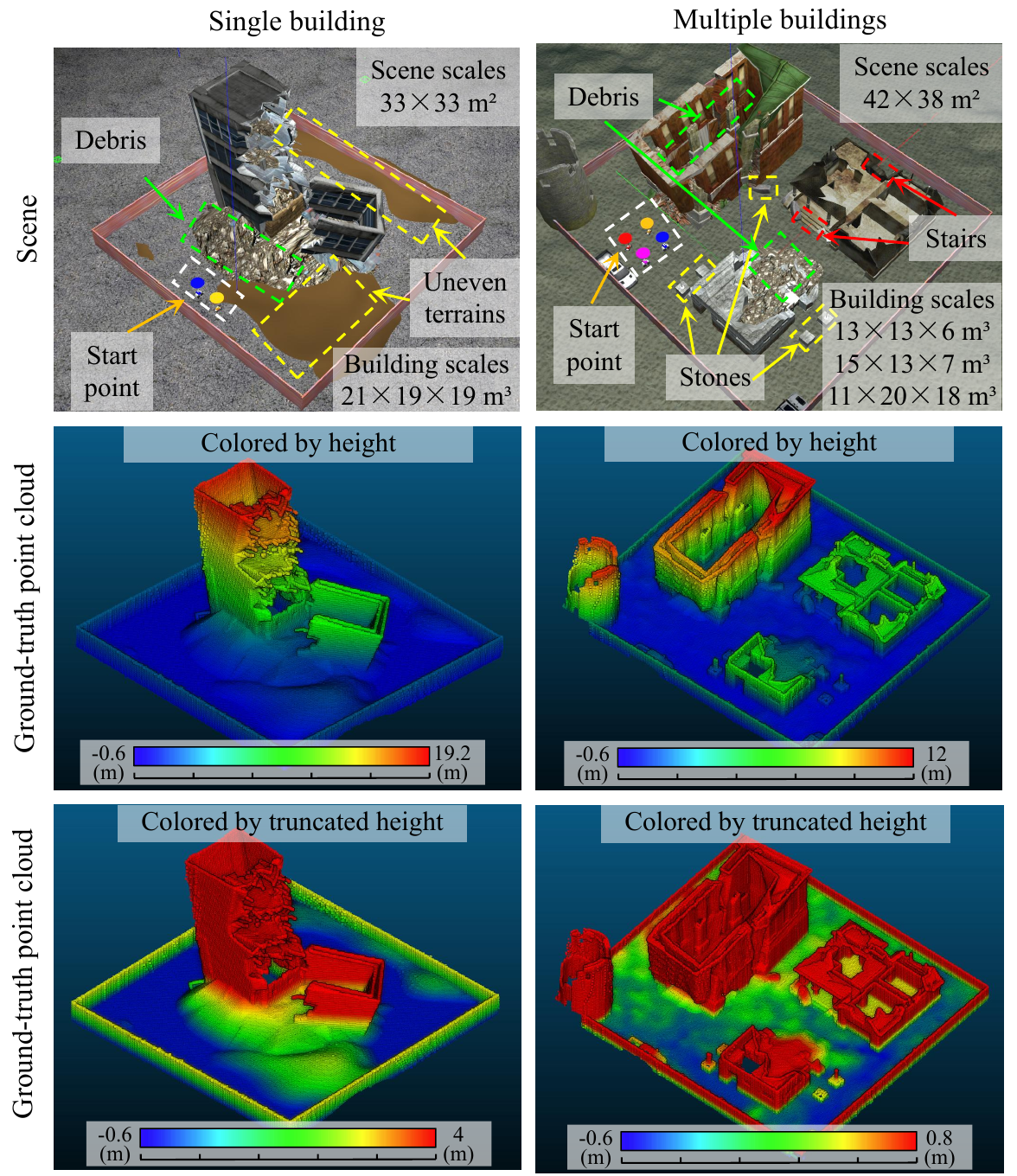}
    \caption{
    	Simulated collapsed building scenes and their ground-truth point clouds.
    	The first row demonstrates the simulated scenes. 
    	The second row shows the ground-truth point clouds colored by height to highlight building surfaces, while the third row depicts the ground-truth point clouds colored by truncated height to emphasize uneven terrain.  
    }
    \label{Fig.4.9.Scene}
\end{figure}

Simulated complex scenes with single or multiple collapsed buildings were constructed to evaluate the proposed and three comparative approaches. 
As shown in Fig. \ref{Fig.4.9.Scene}, the first row demonstrates the simulated scenes and the initial positions of the robots (white dashed boxes), while the second and third rows display the ground-truth point clouds, with colors indicating different height ranges to distinguish building surfaces from uneven terrain.
The single-building scene covers approximately 1,089 m² and has a building volume of about 4,548 m³. 
The building consists of a tall main structure and a short annex that contains indoor spaces accessible only by traversing the debris area (green dashed box). Complete coverage of this environment requires the complementary capabilities of heterogeneous robots. 
The multi-building scene spans approximately 1,554 m² and contains three main buildings with volumes of roughly 1,014 m³, 1,365 m³, and 3,960 m³, respectively. The building interiors contain debris and rooms, while the surrounding areas feature rugged terrain, including stairs and rubble, as well as numerous obstacles like stones and vehicles.  The scene layouts challenge the robots’ autonomous and collaborative capabilities in exploring indoor–outdoor environments.

During exploration, collaborative planning generates a sequence of task views, which are sequentially set as motion-planning goals to capture new point clouds for map updates. To balance map freshness and computational efficiency, a replanning-triggering strategy similar to that in \cite{dong2019multi} was employed, in which replanning is triggered only when the current route times out, i.e. over $\eta_t$, or when any robot finishes its assigned route. 
Furthermore, when one robot enters another's FoV, artifacts can be introduced into the observed point clouds, degrading subsequent mapping and planning. To mitigate this issue, real-time positions were shared among robots to generate 3D bounding boxes based on each robot’s physical dimensions, and point clouds within these boxes were removed from each robot's current sensor frame. Without loss of generality, deep learning-based object detectors can be introduced to autonomously recognize teammate robots, further reducing reliance on communication.

For the single-building scene, a heterogeneous team comprising a SLUGCAT and a MR1000 robot was employed. For the multi-building scene, a team of two SLUGCAT and two MR1000 robots was utilized. The resulting maps and trajectories are visualized in Fig. \ref{fig:11_SingleBuildingPaths} and Fig. \ref{fig:13_MultiBuildingPaths}, respectively, where colors denote heights in the maps and distinct trajectory curves indicate individual robot paths. 
As shown in Fig. \ref{fig:10_SingleBuildingMetrics} and Fig. \ref{fig:12_MultiBuildingMetrics}, exploration performance was evaluated in both single-building and multi-building scenes using multiple key metrics. 
\emph{Reconstruction} completeness is defined as the ratio of the mapped area to the total region of the ground-truth model, reflecting overall coverage progress.
\emph{Exploration time} denotes the total duration of exploration. 
\emph{Cumulative completeness} is the sum of the completeness of each robot, which quantifies inter-robot map overlap. A lower cumulative completeness indicates less overlap and thus more reasonable exploration planning.
\emph{Single-round planning time} refers to the total computational time required for task-view generation, task assignment, and exploration planning, given the current high-level map. 
\emph{Cumulative trajectory length} is the sum of the path lengths traveled by all robots throughout the exploration. A shorter cumulative trajectory implies fewer unnecessary movements and energy costs.

\subsubsection{Exploration Efficiency and Reconstruction Completeness}

\begin{figure}[tpb]
\centering
\includegraphics[width=1\linewidth]{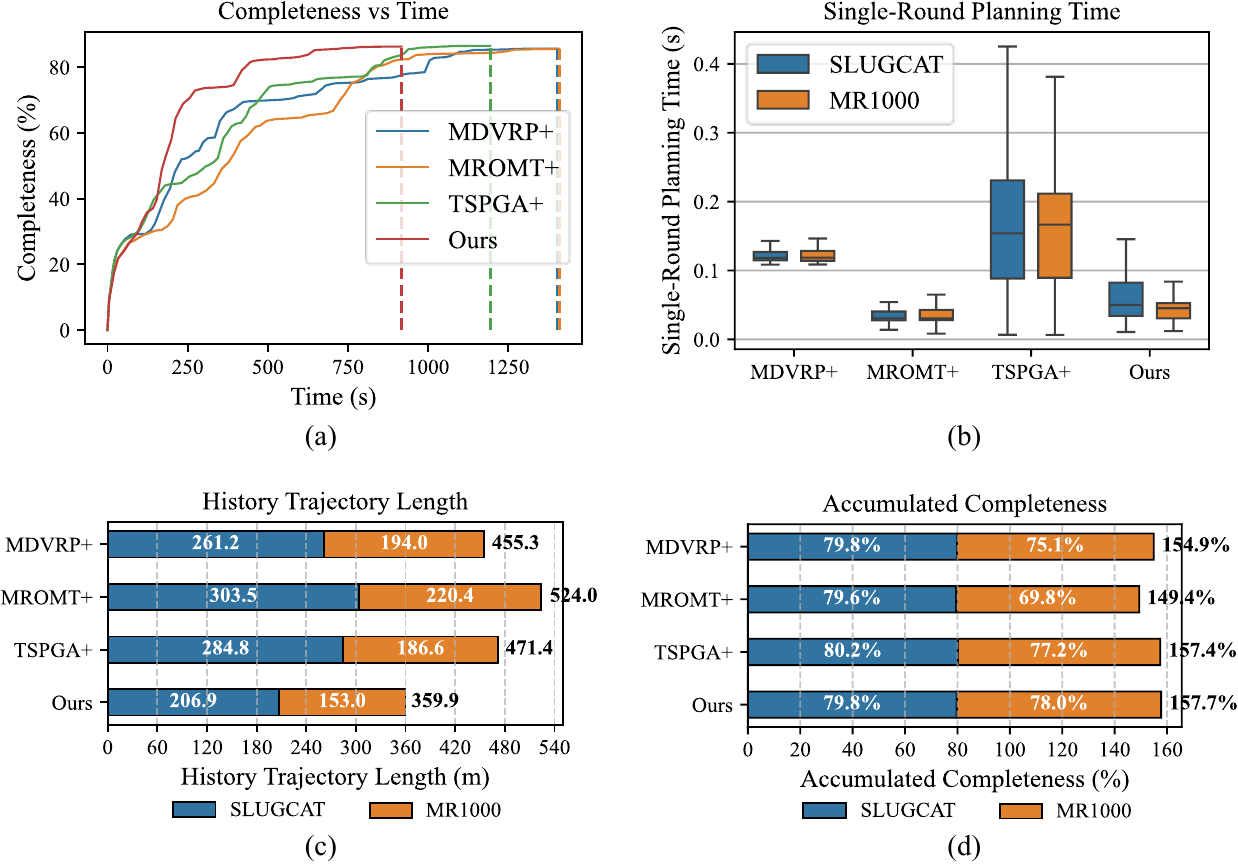}
\caption {Metric results of the four approaches in the simulated single building scene.
	(a) Completeness versus time. 
	(b) Single-round planning time.
	(c) History-trajectory length. 
	(d) Accumulated completeness. 
}
\label{fig:10_SingleBuildingMetrics}
\end{figure}
\begin{figure}[tpb]
\centering
\includegraphics[width=0.42\textwidth]{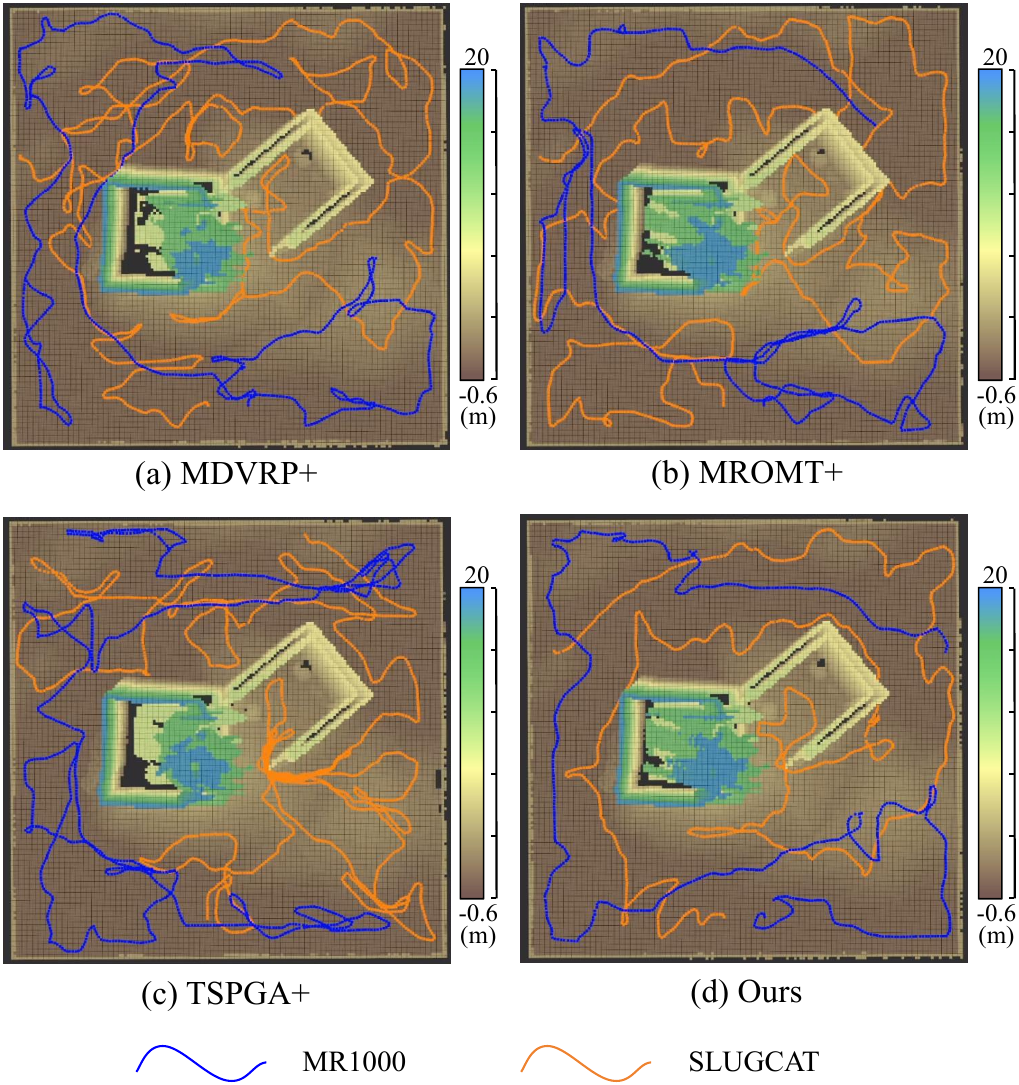}
\caption {Top view of reconstructed maps and exploration trajectories of the four approaches (a)-(d) in the simulated single-building scene. 
}
\label{fig:11_SingleBuildingPaths}
\end{figure}

\begin{figure}[tpb]
    \centering
    \includegraphics[width=0.509\textwidth]{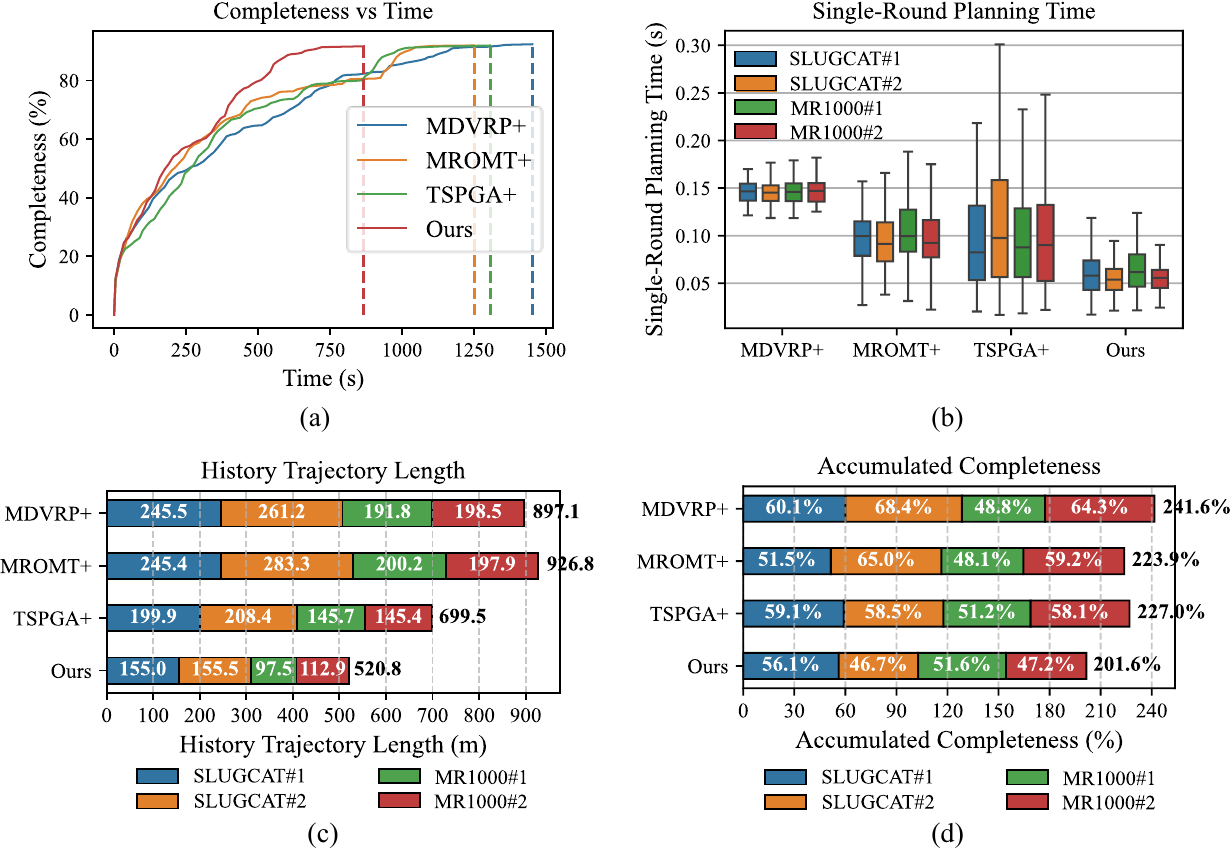}
    \caption{Metric results of the four approaches in the simulated multi-building scene.		
    	(a) Completeness versus time. 
    	(b) Single-round planning time.
    	(c) History-trajectory length. 
    	(d) Accumulated completeness. 
    }
    \label{fig:12_MultiBuildingMetrics}
\end{figure}
\begin{figure}[tpb]
    \centering
    \includegraphics[width=0.48\textwidth]{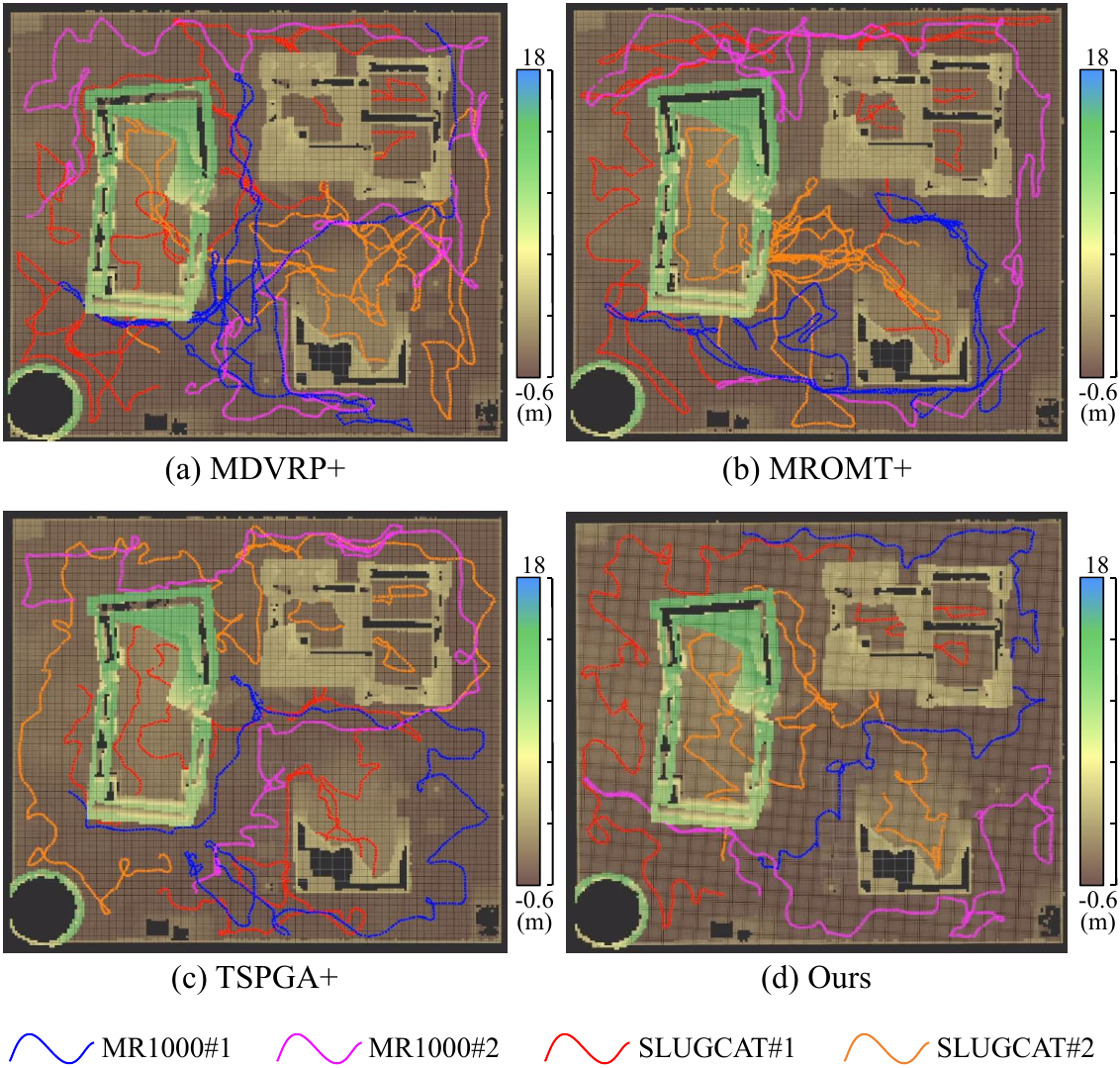}
    \caption{Top view of reconstructed maps and exploration trajectories of four approaches (a)-(d) in the simulated multi-building scene. }
    \label{fig:13_MultiBuildingPaths}
\end{figure}

As for the single-building scene, the proposed approach achieved higher reconstruction completeness than comparative approaches (see Fig. \ref{fig:10_SingleBuildingMetrics}(a)), reducing exploration time by 23.16\%$\sim$34.62\% while producing shorter trajectories (see Fig. \ref{fig:10_SingleBuildingMetrics}(c)). 
In terms of the multi-building scene with complex layouts and increased robots, the advantages of our approach became more pronounced (see Figs. \ref{fig:12_MultiBuildingMetrics}(a) and (c)), with exploration time reduced by 30.68\%$\sim$40.31\%. 
The TSPGA+ and MDVRP+ approaches exhibited rapid volume coverage at the initial stage. Still, they left numerous incomplete surfaces, resulting in extensive detours during the later stages (see Figs. \ref{fig:11_SingleBuildingPaths}(a), \ref{fig:11_SingleBuildingPaths}(c), \ref{fig:13_MultiBuildingPaths}(a), and  \ref{fig:13_MultiBuildingPaths}(c)). 
The TSPGA+ approach was constrained by its greedy sequential assignment strategy, which focuses solely on locally optimal choices and fails to ensure global optimality. Moreover, due to the lack of view merging, the two approaches produced more redundant views for ISE supervoxels, resulting in inefficient exploration routes. 
Although the MDVRP+ approach employed a high-performance solver for task assignment, the limited computational time budget prevented it from obtaining optimal solutions with numerous views, and the solver's inherent randomness further led to frequent route switching and back-and-forth motions. 
Since the MROMT+ approach incorporated view merging, its clustering-based task assignment tended to guide robots toward the regions with denser views. Still, it failed to improve exploration efficiency for two main reasons. On one hand, some views were left during exploration, requiring robots to return later. On the other hand, frequent switching occurs between clusters with similar costs, producing severe detours and the longest trajectories among all approaches (Figs. \ref{fig:10_SingleBuildingMetrics}(c), \ref{fig:12_MultiBuildingMetrics} (c), \ref{fig:11_SingleBuildingPaths} (b), and \ref{fig:13_MultiBuildingPaths}(b)).

In contrast, our approach effectively simplified the assignment problem through task-view clustering and incorporated unified reference routes as constraints into the task assignment, thereby ensuring rational assignment and route consistency for robots. Furthermore, the incorporation of minimal-view-set sampling substantially reduced view redundancy and improved exploration efficiency. As shown in Figs. \ref{fig:11_SingleBuildingPaths}(d) and \ref{fig:13_MultiBuildingPaths}(d), our approach produced dispersed exploration trajectories with less overlap among robots, resulting in lower cumulative completeness in the multi-building scene (see Fig. \ref{fig:12_MultiBuildingMetrics}(d)). 
Conversely, the other approaches exhibited inferior task-assignment consistency, leading to more scanning overlap and, consequently, higher cumulative completeness. In the single-building scene, fully reconstructing the structures requires both robots to circumnavigate the building to reach task views that demand a long observation distance and high traversal capability. Therefore, the cumulative completeness of all approaches is close to each other (see Fig. \ref{fig:10_SingleBuildingMetrics}(d)).

\subsubsection{Computation Runtime}

\begin{table*}[b]
\centering
\caption{Communication-data volumes of four approaches in the multi-building exploration task}
\label{tab:4-4-comm-multi-building-comparison}
\begin{tabular}{@{}|c|cc|cc|cc|cc|cc|@{}}
	\hline
	\multirow{4}{*}{\textbf{\begin{tabular}[c]{@{}c@{}}Approach\end{tabular}}} & \multicolumn{2}{c|}{\textcolor{black}{\textbf{Pose}}} & \multicolumn{2}{c|}{\textbf{Incremental Map}} & \multicolumn{2}{c|}{\textbf{Executable Path}} & \multicolumn{2}{c|}{\textbf{Exploration Route}} & \multicolumn{2}{c|}{\textbf{Total}} \\ \cline{2-11} 
	& \multicolumn{1}{c|}{\textbf{\begin{tabular}[c]{@{}c@{}}Average\\ Rate \\ (KB/s)\end{tabular}}} & \textbf{\begin{tabular}[c]{@{}c@{}}Cumulation \\ (KB)\end{tabular}} & \multicolumn{1}{c|}{\textbf{\begin{tabular}[c]{@{}c@{}}Average \\Rate \\ (KB/s)\end{tabular}}} & \textbf{\begin{tabular}[c]{@{}c@{}}Cumulation \\ (KB)\end{tabular}} & \multicolumn{1}{c|}{\textbf{\begin{tabular}[c]{@{}c@{}}Average \\Rate \\ (KB/s)\end{tabular}}} & \textbf{\begin{tabular}[c]{@{}c@{}}Cumulation \\ (KB)\end{tabular}} & \multicolumn{1}{c|}{\textbf{\begin{tabular}[c]{@{}c@{}}Average \\Rate \\ (KB/s)\end{tabular}}} & \textbf{\begin{tabular}[c]{@{}c@{}}Cumulation \\ (KB)\end{tabular}} & \multicolumn{1}{c|}{\textbf{\begin{tabular}[c]{@{}c@{}}Average \\Rate \\ (KB/s)\end{tabular}}} & \textbf{\begin{tabular}[c]{@{}c@{}}Cumulation \\ (KB)\end{tabular}} \\ \hline
	MDVRP+ & \multicolumn{1}{c|}{\textbf{0.012}} & 14.592 & \multicolumn{1}{c|}{1.162} & 1413.417 & \multicolumn{1}{c|}{0.077} & 94.038 & \multicolumn{1}{c|}{0.075} & 91.800 & \multicolumn{1}{c|}{2.327} & 2829.847 \\ \hline
	MROMT+ & \multicolumn{1}{c|}{\textbf{0.012}} & 14.436 & \multicolumn{1}{c|}{1.098} & 1320.476 & \multicolumn{1}{c|}{0.078} & 94.299 & \multicolumn{1}{c|}{\textbf{0.000}} & \textbf{0.000} & \multicolumn{1}{c|}{2.188} & 2632.211 \\ \hline
	TSPGA+ & \multicolumn{1}{c|}{\textbf{0.012}} & 14.412 & \multicolumn{1}{c|}{\textbf{0.787}} & 944.872 & \multicolumn{1}{c|}{\textbf{0.064}} & 76.638 & \multicolumn{1}{c|}{\textbf{0.000}} & \textbf{0.000} & \multicolumn{1}{c|}{\textbf{1.863}} & 2236.922 \\ \hline
	Ours & \multicolumn{1}{c|}{\textbf{0.012}} & \textbf{10.404} & \multicolumn{1}{c|}{0.989} & \textbf{857.405} & \multicolumn{1}{c|}{0.080} & \textbf{69.360} & \multicolumn{1}{c|}{0.016} & 13.674 & \multicolumn{1}{c|}{2.097} & \textbf{1817.843} \\ \hline
\end{tabular} 
\end{table*}

\begin{figure}[t]
\centering
\includegraphics[width=0.48\textwidth]{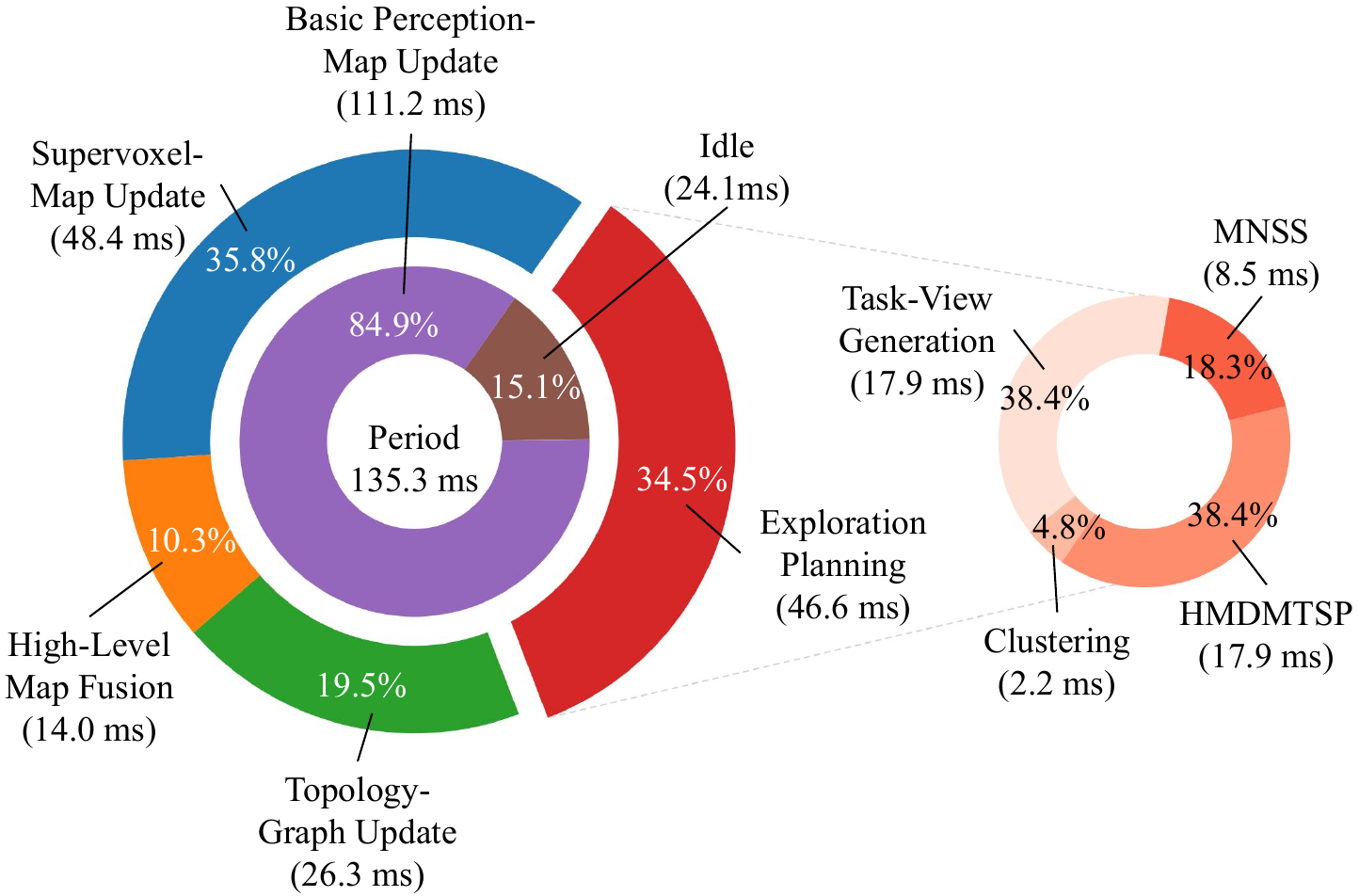}
\caption{Computation runtime of each component of the proposed approach in the simulated multi-building scene. }
\label{Fig.4.14.PieChart}
\end{figure}

The computation runtime of the proposed framework is detailed in Fig. \ref{Fig.4.14.PieChart}, which reports the average runtime of major modules evaluated in the simulated multi-building scene. 
The left pie chart illustrates the parallel and serial modules with two concentric rings. The outer ring indicates the runtime proportions of four serial modules: supervoxel-map update, high-level map fusion, topology-graph update, and exploration planning, with a period of 135.3 ms. The inner ring presents the basic perception-map update, which runs in parallel and marginally faster than the outer-ring modules. 
The right pie chart further decomposes the exploration-planning module into its sub-components, including task-view generation, task-view clustering, HMDMTSP solving, and minimal view-set sampling. 
The basic perception-map update refers to integrating each incoming LiDAR point-cloud into the implicit map while computing traversal and observation metrics. 
The supervoxel map update involves the incremental supervoxel extractions from the basic perception map. The topology-graph update means to refresh the topology graph based on the extracted supervoxels.
High-level map fusion is to integrate the supervoxel map and topology graph increments received from other robots. 

As shown in Fig. \ref{Fig.4.14.PieChart}, the basic perception-map update dominates the overall runtime, primarily due to the substantial computation required to evaluate the traversability for multiple robot species. 
The high-level map update, \textcolor{black}{comprising high-level map fusion (14.0 ms) and topology-graph update (26.3 ms)}, accounted for nearly 30\% of the total period. 
Within exploration planning, task-view generation and HMDMTSP solving accounted for the majority of the computational overhead, owing to the extensive ray-casting operations in the former and the large number of iterations required by the latter. 
The clustering component contributed only a tiny fraction of the period. Overall, the complete framework operated at an average frequency of 7.39 Hz (period 135.3 ms), providing real-time support for online exploration planning.

\subsubsection{Communication-Data Volume}

The communication-data volumes were evaluated in the multi-building exploration task, including \textcolor{black}{poses}, maps, executable paths, and exploration routes published by each robot. 
To ensure planning effectiveness and communication efficiency, all approaches adopted our map representation and incremental transmission strategy. 
The communication-data volume of each robot is first summed and averaged by the robot number to obtain the cumulative data volume, which is then divided by the total exploration time to obtain the average data volume transmitted per second. 
\textcolor{black}{The pose data} consist of each robot’s real-time coordinates. The map data comprises incrementally published supervoxels and the topology graph. 
The \textcolor{black}{pose}, map, and executable path data were broadcast once per second. The exploration-route data were transmitted only by MDVRP+ and our approach after each planning round: the former shares task-view sequences as initial solutions for the next planning round, whereas the latter shares task-cluster sequences.  

As shown in Table \ref{tab:4-4-comm-multi-building-comparison}, our approach achieved the lowest cumulative map-transmission volume. In contrast, the others suffered from prolonged exploration among robots, resulting in higher cumulative map-transmission volumes. 
The TSPGA+ approach exhibited a relatively low average map transmission rate because only private tasks remained in the late stages of exploration, and some robots were assigned no task. These idling robots published no updated map or executable path, lowering the transmission volume. 
Our approach transmitted less data for executable paths than the MDVRP+ approach, since MDVRP+ transmits all task-view coordinates, whereas ours transmits only task-cluster center coordinates. 
The MDVRP+ approach resulted in a longer exploration time and more planning rounds, leading to a larger executable path-transmission volume.
Overall, owing to higher exploration efficiency, our approach yielded the lowest cumulative communication-data volume, reducing by 18.734\% to 35.762\% compared to others. Additionally,  the communication-data volumes of three map transmission strategies were compared (see Table \ref{tab:4-5-comm-multi-building-map-publish}). The incremental high-level map transmission dramatically saved communication overhead, achieving a 97.816\% reduction compared with full high-level map transmission and a 99.816\% reduction compared with raw basic perception-map transmission. The results demonstrate that our high-level map simplifies map information while preserving essential details, enabling efficient exploration planning with minimal communication burden.

\begin{table}[t]
\centering
\caption{Communication-data volume of different map-transmission strategies in the multi-building exploration task}
\label{tab:4-5-comm-multi-building-map-publish}
\begin{tabular}{|c|c|c|}
	\hline
	\textbf{Map-Transmission Strategy}  & \textbf{Avg. Rate (KB/s)} & \textbf{Cumulative (KB)} \\ \hline
	\begin{tabular}[c]{@{}c@{}}Incremental High-Level\\ Map Transmission\end{tabular} & 0.989 & 857.405 \\ \hline
	\begin{tabular}[c]{@{}c@{}}Full High-Level\\ Map Transmission\end{tabular} & 45.288 & 39,265.014 \\ \hline
	\begin{tabular}[c]{@{}c@{}}Full Basic Perception-\\ Map Transmission\end{tabular} & 537.975 & 466,424.097 \\ \hline
\end{tabular}
\end{table}

\subsection{Ablation Study}\label{sec:ablationStudy}
\subsubsection{Ablated Variants}
An ablation study was conducted in the simulated multi-building scene to validate each component of our framework, including heterogeneous capability adaptation, task-view clustering, MNSS, and task-cluster assignment. The ablated variants are denoted as \emph{NoHetero}, \emph{NoCluster}, \emph{NoMNSS}, and \emph{NoAssign}, while the complete version is denoted as \emph{Full}. 
Note that the heterogeneous capability adaptation refers to the proposed mechanisms in this paper to support robot heterogeneity, including computing traversability on the basic perception map for different robot species (see Sec. \ref{sec:map_representation_fusion}),
estimating traversal-capability requirements in the high-level map (see Sec. \ref{sec:supervoxel_map}),
evaluating observation-capability requirements of task views according to sensor differences (see Sec. \ref{sec:task_view_generation}), and
performing task assignment that explicitly accounts for robot heterogeneity (see Sec. \ref{sec:hdmtsp}). 

The \emph{NoHetero} variant disabled heterogeneous capability adaptation and assumed that all robots have identical observation and traversal capabilities, thereby reducing the assignment problem to a homogeneous multi-robot case. 
The \emph{NoCluster} variant removed the greedy sequential clustering of task views (Sec. \ref{sec:view_clustering}). Task views were used directly as assignment units in the HMDMTSP, and the MNSS didn't work accordingly.
The \emph{NoMNSS} variant eliminated the task-view simplification during exploration-tour generation  (Sec. \ref{sec:exploration_tour_generation}), where a constrained TSP was solved directly over all views within each cluster, with additional precedence constraints imposed according to the intra-cluster order. 
The \emph{NoAssign} variant discard both task-view clustering and the HMDMTSP solving (Sec. \ref{sec:hdmtsp}). Instead, a dynamic Voronoi partitioning method \cite{dong2024fast} was adopted to assign task views, where distances from each task view to all robots are computed on the traversal-topology graph, with MNSS retained for view reduction. 
Quantitative results of the variants are presented in Fig.  \ref{Fig.4.15.AblationMetrics}. 
Top-down views of reconstructed maps and exploration trajectories are shown in Fig. \ref{Fig.4.16.AblationPaths}, where maps are colored by height, and robots' trajectories are rendered with distinct colors.

\subsubsection{Exploration Efficiency and Reconstruction Completeness}
\begin{figure}[t]
	\centering
	\includegraphics[width=0.5\textwidth]{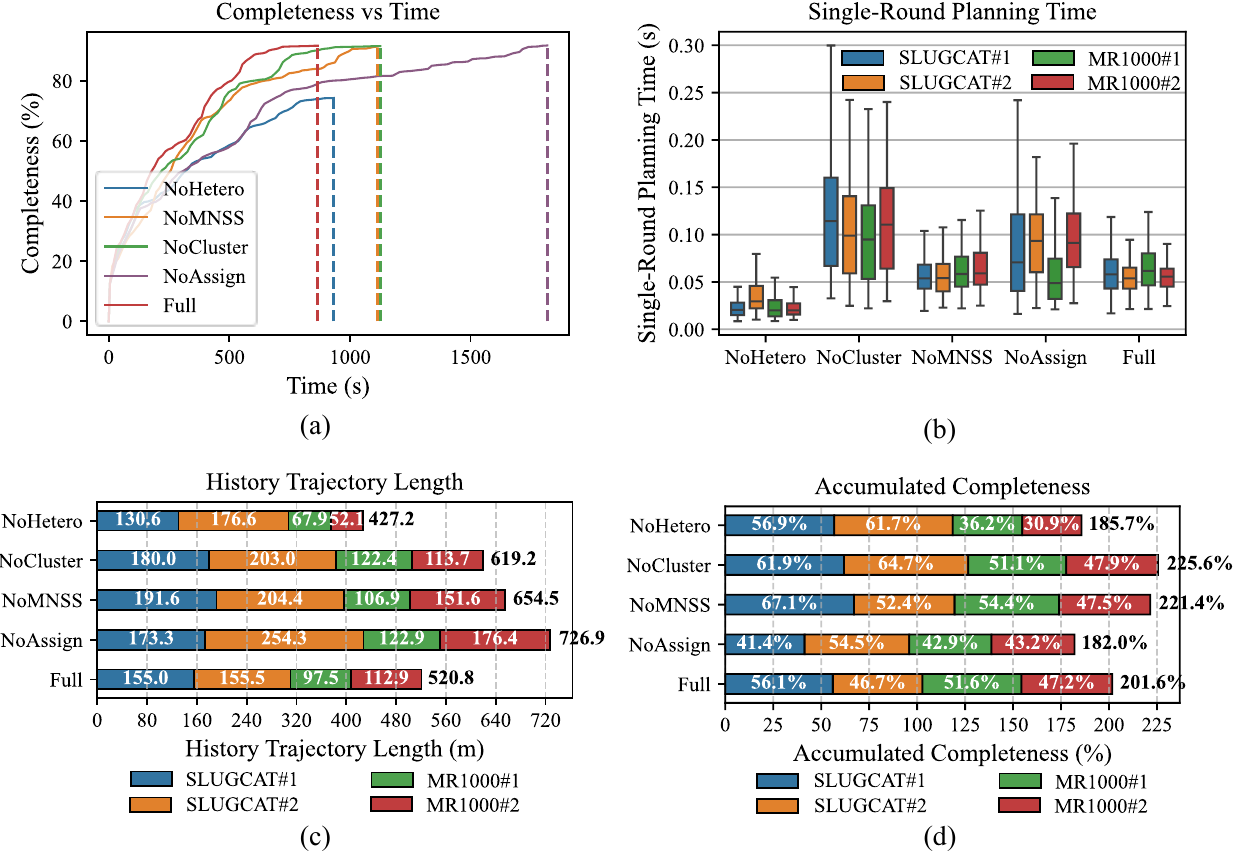}
	\caption {Metric results of the ablated approaches in the simulated multi-building scene. 		
		(a) Completeness versus time. 
		(b) Single-round planning time.
		(c) History-trajectory length. 
		(d) Accumulated completeness. }
	\label{Fig.4.15.AblationMetrics}
\end{figure}

\begin{figure}[t]
	\centering
	\includegraphics[width=0.47\textwidth]{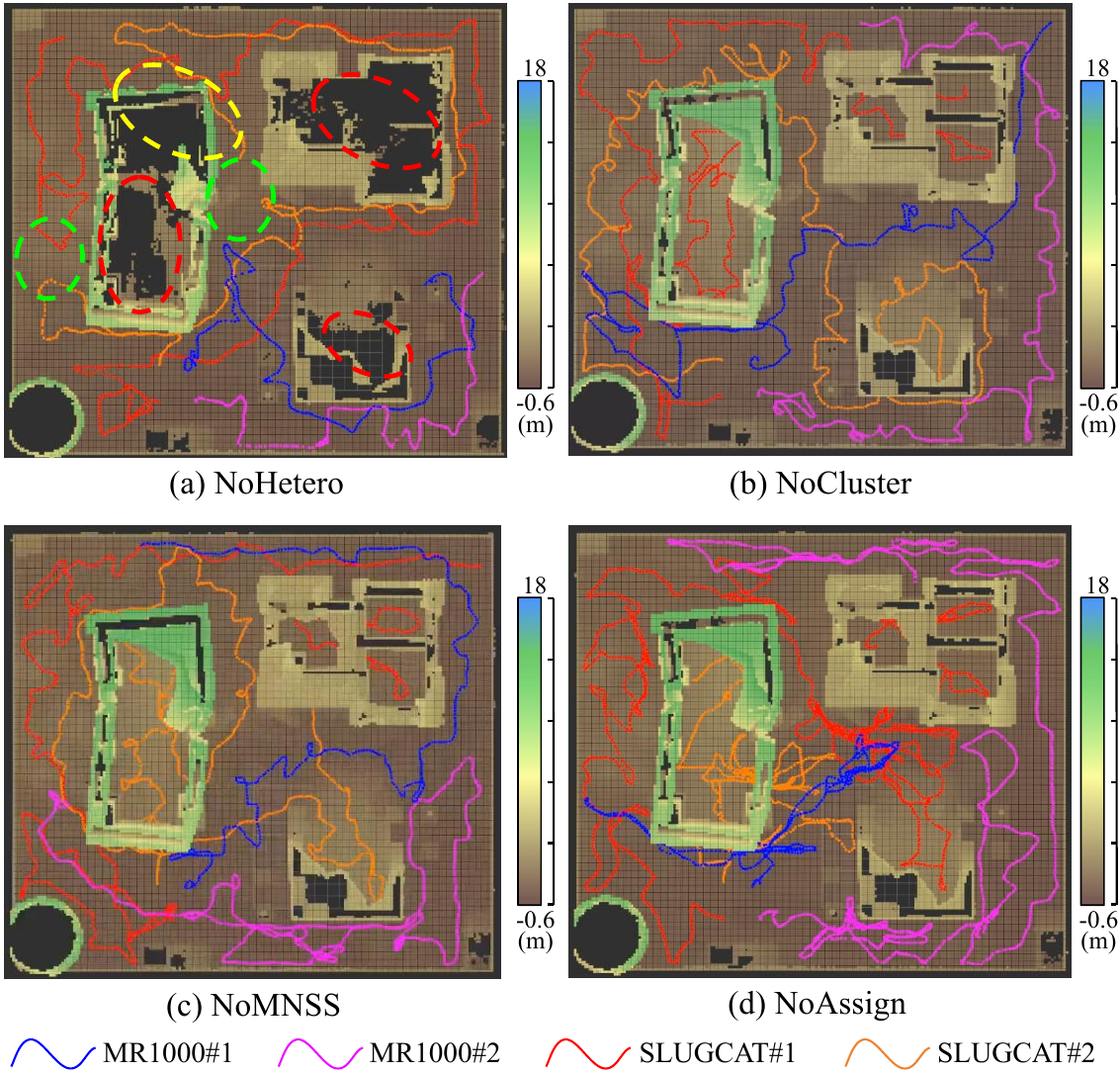}
	\caption{Top view of reconstructed maps and exploration trajectories of the ablated variants (a)-(d) in the simulated multi-building scene. 
	}
	\label{Fig.4.16.AblationPaths}
\end{figure}

The \emph{NoHetero} variant impaired the full release of robot’s traversal and observation potentials. 
On one hand, robots could no longer traverse rough terrains or enter building interiors, leaving large indoor regions unexplored (red dashed circles in Fig. \ref{Fig.4.16.AblationPaths}(a)). 
On the other hand, reduced sensing range caused missing rooftops in the reconstructed map (yellow dashed circles in Fig.  \ref{Fig.4.16.AblationPaths}(a)). 
These missed areas and measurement gaps directly decreased the reconstruction completeness  (see  Fig.  \ref{Fig.4.15.AblationMetrics}(a)). 
Moreover, the weaker traversability assumptions prevented robots from crossing rough terrains (green circles in Fig.  \ref{Fig.4.16.AblationPaths}(a)), resulting in increased path lengths of SLUGCAT robots (see Fig. \ref{Fig.4.15.AblationMetrics}(d)) and reduced overall exploration efficiency. Nevertheless, thanks to the remaining components of our framework, the trajectories remained clean with minimal overlap, comparable to those of the full approach (see Figs. \ref{Fig.4.16.AblationPaths}(a) and \ref{fig:13_MultiBuildingPaths}(d)).
Meanwhile, the \emph{NoHetero} variant yielded the smallest explored area and the shortest trajectory length among all variants (see Fig. \ref{Fig.4.15.AblationMetrics}(c)), whereas the other variants produced longer trajectories than the full approach due to the removal of crucial exploration-planning components.

The \emph{NoMNSS} and \emph{NoCluster} variants removed MNSS and task-view clustering, leading to redundant observations and back-and-forth motions (see Figs. \ref{Fig.4.16.AblationPaths}(b) and (c)), separately. Therefore, their exploration time was prolonged (see Fig. \ref{Fig.4.15.AblationMetrics}(a)), and their trajectory lengths were increased (\ref{Fig.4.15.AblationMetrics}(c)). 
However, their exploration efficiency was not severely degraded by the presence of reference-route constraints. 
In contrast, the \emph{NoAssign} variant removed task-cluster assignment and incurred the largest efficiency loss among all ablations (see Fig. \ref{Fig.4.15.AblationMetrics}(a)). Without the reference-route constraint, relying solely on dynamic Voronoi partitioning caused route jumps between consecutive planning rounds, resulting in substantial route overlaps (see Fig. \ref{Fig.4.16.AblationPaths}(d)). Consequently, both history-trajectory length and exploration time were significantly higher than those of the full approach (see Figs.  \ref{Fig.4.15.AblationMetrics}(a) and (c)).
The \emph{NoAssign} variant exhibited the lowest accumulated completeness (see Fig. \ref{Fig.4.15.AblationMetrics}(d)) due to the severe back-and-forth motions of MR1000$\#1$ and SLUGCAT$\#2$ robots, confining their exploration to very limited regions (see Fig. \ref{Fig.4.16.AblationPaths}(d)). 
The \emph{NoHetero} variant ranked second-lowest in accumulated completeness, since the removed functions significantly reduced the explorable areas. 
Specially, the \emph{NoCluster} and \emph{NoMNSS} variants yielded higher accumulated completeness than the full approach, because their detours increased overlaps among the regions explored by different robots.

\subsubsection{Single-Round Planning Time}
The \emph{NoHetero} variant achieved the lowest single-round planning time (see Fig. \ref{Fig.4.15.AblationMetrics}(b)). Its assumption on shorter observation range reduced candidate views and visibility checks. Meanwhile, its assumption on lower traversal capability decreased task views. 
The \emph{NoMNSS} variant varied little on computational efficiency, because the dominant cost—TSP solving over all assigned views per robot—remains comparable to that in the full approach.
The \emph{NoClustering} variant significantly increased the single-round planning time, since directly solving HMDMTSP across all task views is considerably more expensive. 
Although the \emph{NoAssign} variant eliminated HMDMTSP solving, the numerous task views dramatically increased the computational cost of MNSS, severely degrading overall planning efficiency.

The results above demonstrate the rationality and necessity of each module in the proposed framework. Removing any single component negatively affected either exploration effectiveness or planning efficiency, or both. By integrating all modules, our approach achieved comprehensive improvements that would be unattainable by any individual component alone. 

\subsection{Field Experiments}
\begin{figure}[tbp]
\centering
\includegraphics[width=0.48\textwidth]
{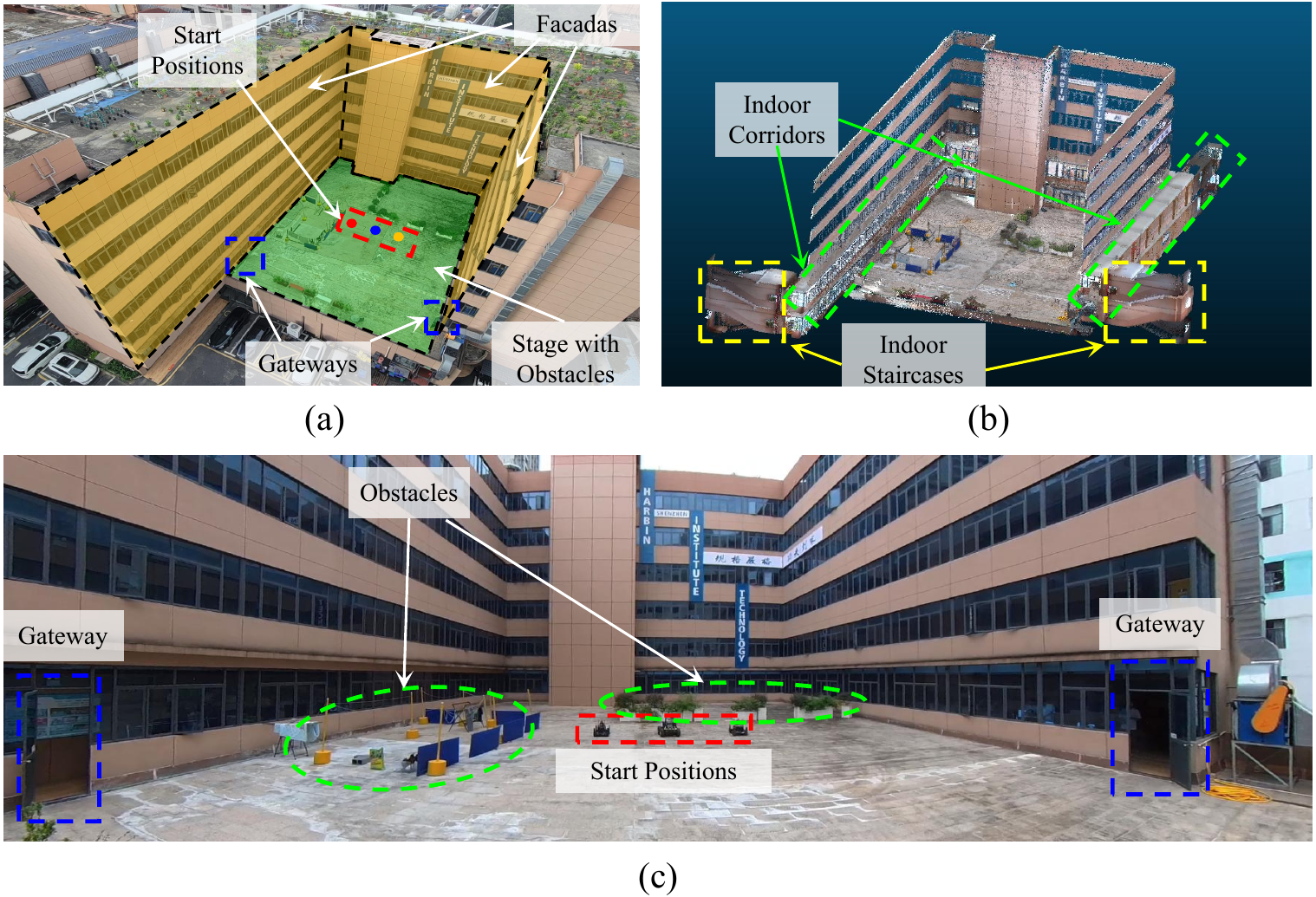}
\caption {The real-world scene and its ground truth point cloud. 
	(a) Overview of the field scene.
	(b) The ground-truth point cloud of the scene.
	(c) The layout of outdoor regions and robot-start positions.
}
\label{Fig.4.17.RealScene}
\end{figure}

\begin{figure}[tbp]
\centering
\includegraphics[width=0.48\textwidth]{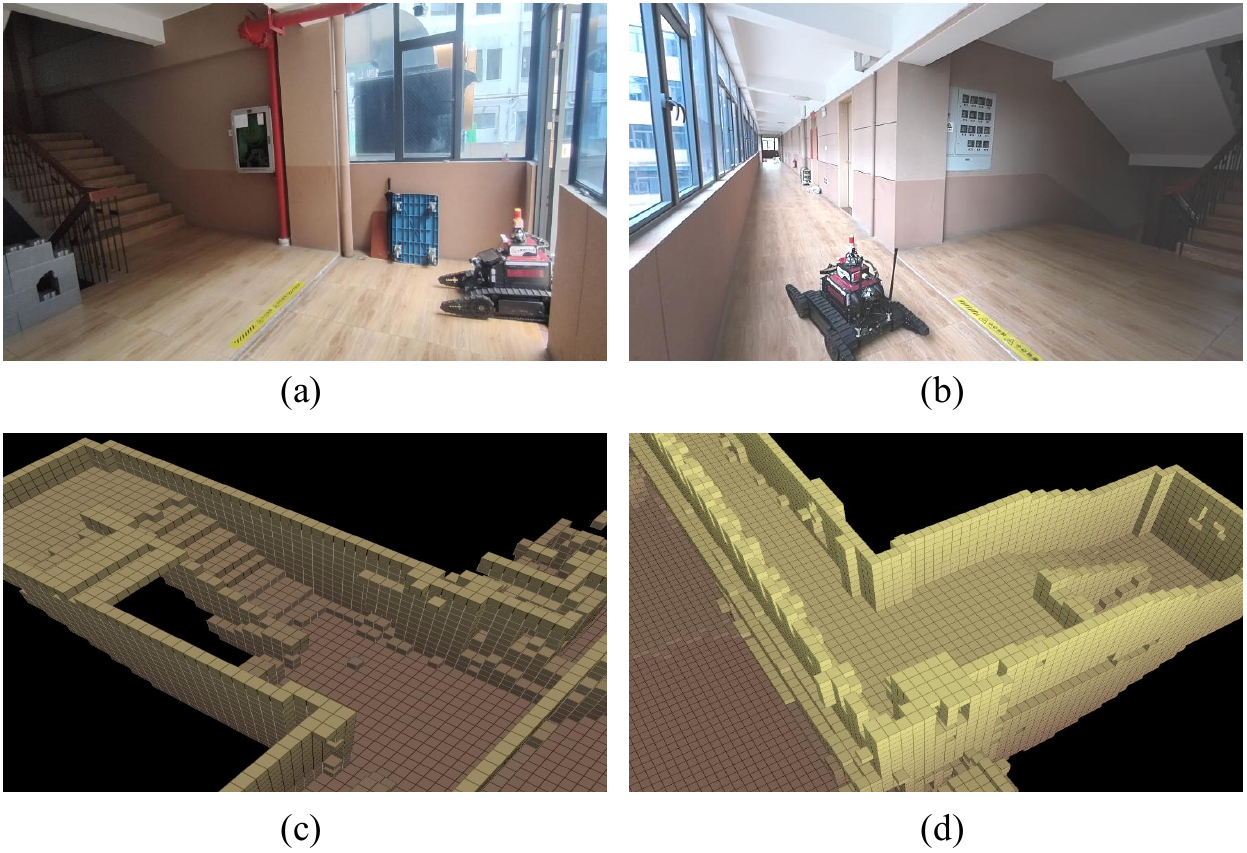}
\caption{The real-world scene and reconstructed maps in different regions. (a) One of the two building entrances near indoor staircases. (b) An indoor corridor near a staircase. (c)(d) The maps corresponding to (a) and (b).  }
\label{Fig.4.18.IndoorSceneMap}
\end{figure}

The field experiment was conducted in a complex indoor-outdoor environment with a total volume of approximately 60,000 m³, as shown in Fig.~\ref{Fig.4.17.RealScene}(a). The scene featured diverse layouts and terrains, challenging the collaborative exploration of heterogeneous multiple robots. 
The scene centered on an outdoor platform cluttered with obstacles and was surrounded by a multi-layer building. 
The platform provided two entrances leading into indoor corridors of the building (see Fig.  \ref{Fig.4.18.IndoorSceneMap}(a)). 
These corridors extended inward and connected to staircases that access higher floors (see Fig. \ref{Fig.4.18.IndoorSceneMap}(b)). 
Therefore, the robots were required to navigate diverse terrains, explore both indoor and outdoor regions, and perform efficient 3D reconstruction. 
The ground-truth point cloud of the entire scene (see Fig.  \ref{Fig.4.17.RealScene}(b)) was manually obtained for evaluation using a high-precision 3D laser scanner\footnote{https://www.feimarobotics.com/zhcn/productDetailSLAM2000}. 

The physical multi-robot system consisted of three heterogeneous robots: one MR1000 robot and two SLUGCAT robots. The robot configurations and sensor suites are detailed in Figs.  \ref{fig:robots}(b) and (d). 
Based on the tight-coupled SLAM framework \cite{fastlio2}, multi-sensor-fused SLAM was developed for each sensor suite. 
For the myopic panoramic sensor suite, point clouds from three LiDARs were motion-compensated using IMU data and registered using calibrated extrinsic parameters to improve frame-to-map motion estimation. 
For the hyperopic gimbal-sensor suite, the primary odometry was estimated using the bottom-mounted MID360 LiDAR. The measured gimbal angles were then combined with the primary odometry to correct the IMU-integrated state estimation of the top-mounted Horizon LiDAR.  
The robots were initially positioned on the outdoor platform, as indicated by the red dashed boxes in Figs. \ref{Fig.4.17.RealScene}(a) and (c), and coordinate frames were aligned through the registration process at the beginning of experiment.

\begin{table*}[b]	
\caption{Results of exploration metrics in the real-world scene}
\label{tab:4-6-field-exp-metrics}
\centering
\begin{tabular}{|c|c|c|c|c|}
	\hline
	\textbf{Robot} & \textbf{\begin{tabular}[c]{@{}c@{}}Trajectory\\ Length (m)\end{tabular}} & \textbf{\begin{tabular}[c]{@{}c@{}}Rescontruction\\ Completeness (\%)\end{tabular}} & \textbf{\begin{tabular}[c]{@{}c@{}}Communication-\\ Data Volume (KB)\end{tabular}} & \textbf{\begin{tabular}[c]{@{}c@{}}Average\\ Communication Rate (KB/s)\end{tabular}} \\ \hline
	SLUGCAT\#1 & 126.350 & 66.625 & 1,526.898 & 1.152 \\ \hline
	SLUGCAT\#2 & 115.846 & 51.798 & 1,484.921 & 1.121 \\ \hline
	MR1000 & 100.241 & 76.297 & 2,606.411 & 1.967 \\ \hline
	Total & 342.437 & 194.721 & 5,618.230 & 4.240 \\ \hline
\end{tabular}
\end{table*}

\begin{figure}[tb]
	\centering
	\includegraphics[width=0.47\textwidth]
	{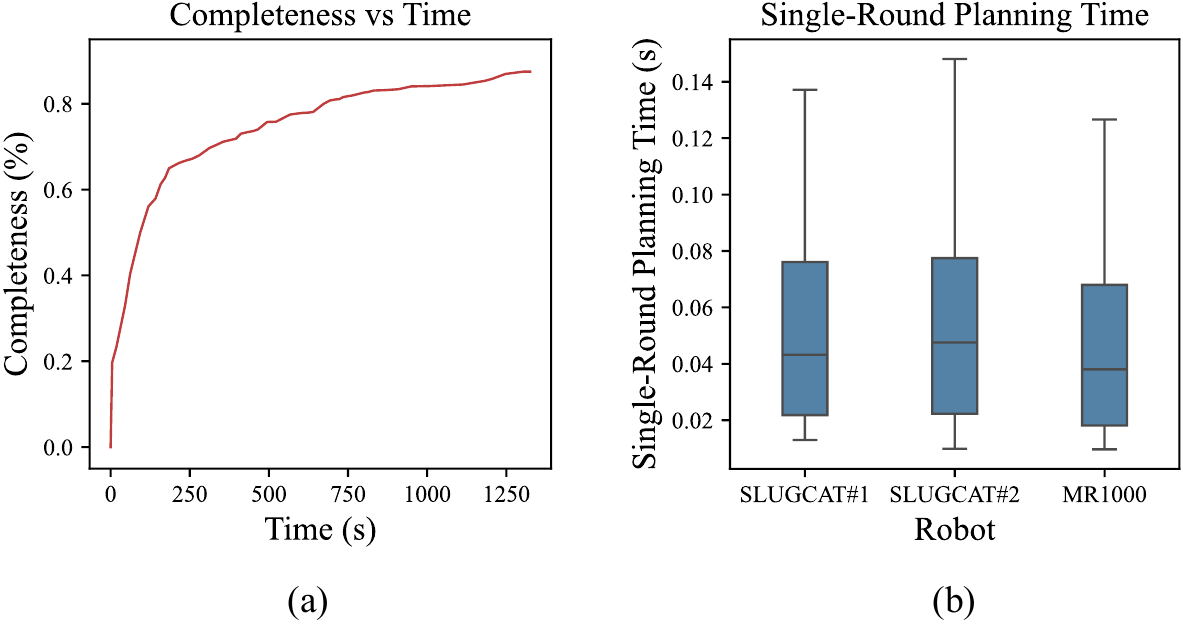}
	\caption{Results of exploration metrics in the real-world scene. (a) Completeness versus time. (b) Single-round planning time of each robot. }
	\label{Fig.4.19.FieldMetrics}
\end{figure}
\begin{figure}[tb]
\centering
\includegraphics[width=0.47\textwidth]
{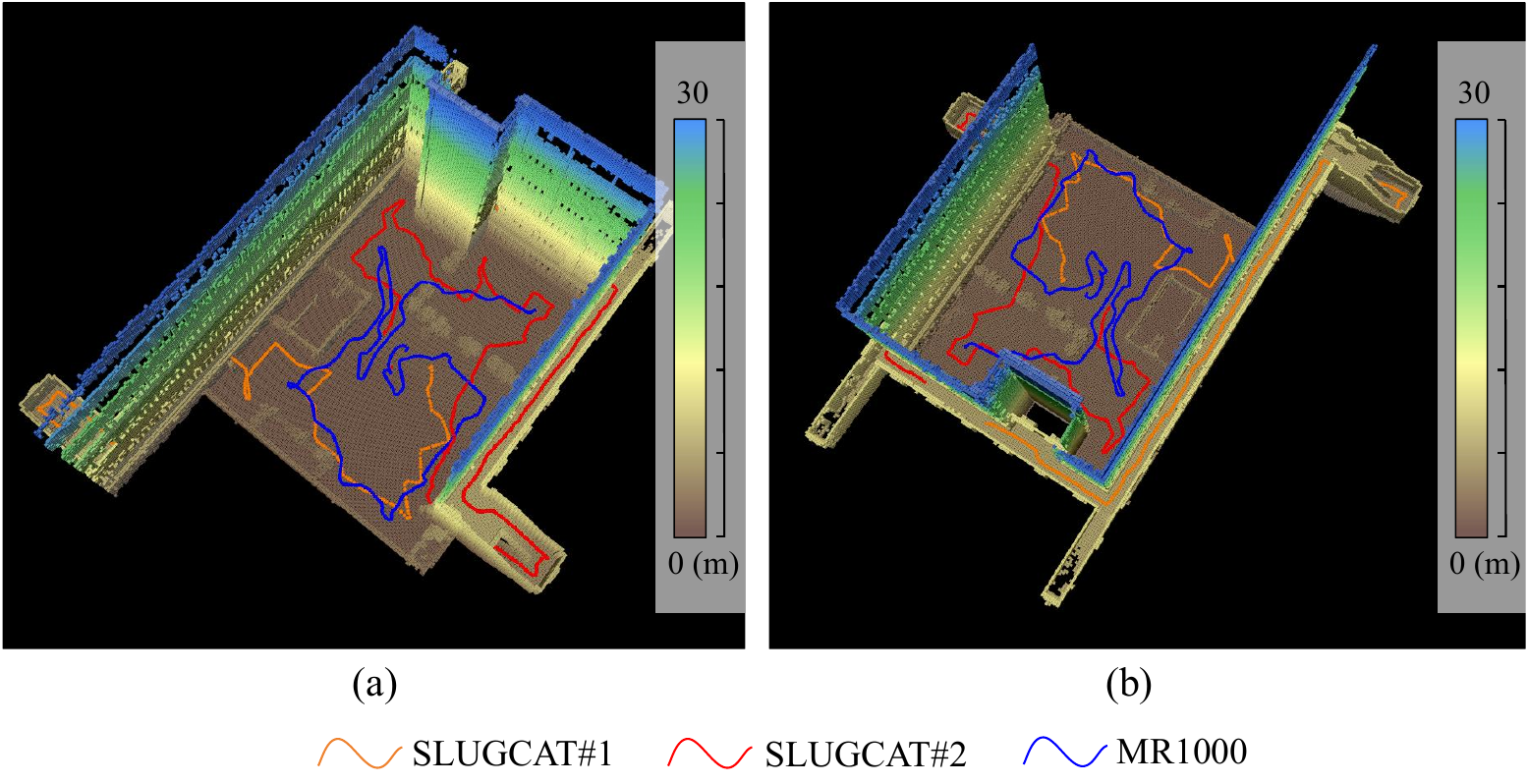}
\caption{Two bird-eye views (a) and (b) of reconstructed map and exploration trajectories in the real-world scene. }
\label{Fig.4.20.FieldPaths}
\end{figure}

Key metrics were recorded during the exploration process. 
As shown in Fig. \ref{Fig.4.19.FieldMetrics}(a), the reconstruction completeness and exploration time demonstrate that the heterogeneous robot team efficiently completes the exploration of both indoor and outdoor areas through effective collaboration. As presented in Fig. \ref{Fig.4.19.FieldMetrics}(b), the single-round planning time for each robot is quite similar, consistent with the results in simulations (see  Figs. \ref{fig:10_SingleBuildingMetrics}(b) and \ref{fig:12_MultiBuildingMetrics}(b)). 
More detailed statistics are provided in Table \ref{tab:4-6-field-exp-metrics}. 
Since the building facades constituted the majority of observable environmental surfaces, the MR1000 robot was dominated by the task views of high-rise facades. Meanwhile, this robot achieved higher reconstruction completeness than the other two, thereby incurring greater communication-data volume and bandwidth usage. 
The actual exploration trajectories are illustrated in  Fig. \ref{Fig.4.20.FieldPaths}. In the initial stage, the two SLUGCAT robots worked alongside the MR1000 robot to rapidly explore the outdoor terrain. Subsequently, they entered the indoor areas through entrances on both sides of the outdoor platform, climbed staircases to upper floors for continued exploration, and \textcolor{black}{finished} at the end of corridor. 
Reconstruction results of the indoor scene are partially shown in  Figs. \ref{Fig.4.18.IndoorSceneMap}(c) and (d), with height truncated at 2.5 m and 5 m, respectively, to avoid visualization occlusion. 
As for trajectory length (see Table \ref{tab:4-6-field-exp-metrics}), the MR1000 robot produced a shorter trajectory, since it primarily scanned the building facades. In contrast, the two SLUGCAT robots explored additional indoor regions, leading to longer trajectories. The exploration process fully exploited the potential of the heterogeneous robots, thereby achieving successful 3D reconstruction of both indoor and outdoor environments.
For more details, please refer to the supplementary video.

\section{Conclusion and Future Work}   \label{sec:conclusion}
This paper proposes a collaborative exploration framework designed for heterogeneous multi-robot teams operating in complex 3D environments. 
First, a basic perception-map representation is introduced by integrating terrain metrics and observation metrics. 
Second, an improved supervoxel segmentation is designed to address mis-segmentation of complex scenes. 
The produced supervoxels are further utilized to construct a high-level abstract map for low-bandwidth communication among multiple robots. 
Furthermore, task views are generated from the high-level map and clustered according to robot capabilities to reduce exploration-planning complexity. 
Subsequently, the task-cluster assignment is formulated as a HMDMTSP with additional constraints between task requirements and robot capabilities, and an improved genetic algorithm is designed to solve it efficiently. 
Redundant views are further removed from task clusters, and efficient exploration routes are generated for robots. 
Additionally, specific strategies are provided to resolve conflicts of executable paths to ensure safety among multiple robots. 
Finally, evaluations are conducted in both simulation and real-world scenes, with comparisons against state-of-the-art approaches. 
The results demonstrate that our approach achieves effective collaborative exploration and outperforms the other approaches across various scenes and robot teams.

Future work will focus on three directions. 
First, incorporating richer information, such as semantic features, into the map to support higher-level decision making. 
Second, developing multimodal map-based exploration strategies for more complex workspace. 
Third, scaling to cross-domain heterogeneous robot teams, such as aerial, amphibious, and other robots.


\bibliographystyle{IEEEtran}
\bibliography{IEEEabrv,reference}

\end{document}